\documentclass[envcountsame,runningheads]{llncs}

\usepackage{graphicx}                   
\usepackage{hyperref}                   

\newtoggle{arxiv}
\toggletrue{arxiv}

\usepackage{ellipsis, ragged2e} 
\usepackage[l2tabu, orthodox]{nag} 

\usepackage[english]{babel}
\usepackage[T1]{fontenc}
\usepackage[utf8]{inputenc}
\usepackage[final]{microtype}

\usepackage{bbm}
\usepackage{proof}
\usepackage{csquotes}
\usepackage{mathtools}
\usepackage{booktabs}
\usepackage{multirow}
\usepackage{afterpage}
\usepackage{xparse}
\usepackage{hyperref}
\usepackage{enumitem}
\usepackage[normalem]{ulem}

\usepackage[table]{xcolor}

\usepackage[capitalise]{cleveref}

\input{preamble/0_macros}
\usepackage{tikz}

\usetikzlibrary{automata,positioning,shapes,calc,fit,arrows,arrows.meta,backgrounds,colorbrewer}
\usepackage{pgfplots}
\usepgfplotslibrary{fillbetween}
\pgfplotsset{compat=1.13}

\tikzstyle{state}+=[minimum size=20pt,inner sep=2pt]
\tikzstyle{action}=[font=\small,inner sep=0pt,outer sep=3pt]
\tikzstyle{actionedge}=[->,draw]
\tikzset{chainarrow/.tip={Stealth[length=3pt]}}
\tikzset{>=chainarrow}

\tikzstyle{system} = [draw,circle,minimum size=1cm]
\tikzstyle{environment} = [draw,diamond,minimum size=1cm]

\tikzstyle{systemltl} = [draw,rectangle,rounded corners=5pt,minimum size=1cm]
\tikzstyle{environmentltl} = [draw,rectangle,minimum size=1cm]

\tikzset{every picture/.append style={
	auto,
	node distance=2cm,
	initial text={},
}}

\pgfplotscreateplotcyclelist{no marks}{%
	loosely dotted, mark=\empty\\
	densely dotted, mark=\empty\\
	solid, mark=\empty\\
	densely dashed, mark=\empty\\
	loosely dashed, mark=\empty\\
	dashdotted, mark=\empty\\
	dashdotdotted, mark=\empty\\
	densely dashdotted, mark=\empty\\%
}

\newcommand{\plotlimit}{1800}
\newcommand{\TO}{3600}
\newcommand{\legendmax}{4600}
\newcommand{\plotmarksize}{1.5pt}
\newcommand{\axislines}{
	\addplot[black,forget plot,update limits=false] coordinates {(0.0000001,0.0000001) (10000,10000)};
	\addplot[black,dashed,forget plot,update limits=false] coordinates {(0.0000001,0.0000002) (4000,8000)};
	\addplot[black,dashed,forget plot,update limits=false] coordinates {(0.0000002,0.0000001) (8000,4000)};
	
	\addplot[gray,dashed,forget plot,update limits=false] coordinates {(0.0000001,\TO) (10000,\TO)};
	\addplot[gray,dashed,forget plot,update limits=false] coordinates {(\TO,0.0000001) (\TO,10000)};
	\addplot[gray,thin,forget plot,update limits=false] coordinates {(\plotlimit,0.0000001) (\plotlimit,\plotlimit)};
	\addplot[gray,thin,forget plot,update limits=false] coordinates {(0.0000001,\plotlimit) (\plotlimit,\plotlimit)};
}
\newcommand{\markincaption}[2]{{\begin{tikzpicture}[baseline=-3pt] \protect\draw[#1,thick,mark size=3pt] plot[mark=#2] (0,0);\end{tikzpicture}}}

\newcommand{\syntunrealplot}{\addplot+[only marks,draw=Dark2-B,mark size=\plotmarksize,mark=star]}
\newcommand{\syntheticplot}{\addplot+[only marks,draw=Dark2-C,mark size=\plotmarksize,mark=asterisk]}
\title{
SemML: Enhancing Automata-Theoretic LTL Synthesis with Machine Learning \thanks{
	This research was funded in part by the DFG project 427755713 GOPro and the MUNI Award in Science and Humanities
	(MUNI/I/1757/2021) of the Grant Agency of Masaryk University.}
}

\author{
	Jan~K\v{r}et{\'i}nsk{\'y}\inst{1,2} \orcidID{0000-0002-8122-2881} \textsuperscript{(\Letter)}
	\and Tobias~Meggendorfer\inst{3} \orcidID{0000-0002-1712-2165}
	\and Maximilian~Prokop\inst{1,2} \orcidID{0009-0008-6512-8693}
	\and Ashkan~Zarkhah\inst{1} \orcidID{0009-0004-9510-5485}
}

\institute{
	Masaryk University, Brno, Czech Republic
	\and
	Technical University of Munich, Munich, Germany
	\and
	Lancaster University Leipzig, Leipzig, Germany
}

\let\llncssubparagraph\subparagraph
\let\subparagraph\paragraph
\usepackage{titlesec}
\let\subparagraph\llncssubparagraph

\titlespacing*{\subsubsection}{0pt}{1ex plus 0.5ex minus 0ex}{*1}
\titlespacing*{\paragraph}{0pt}{0.75ex plus 0.5ex minus 0ex}{*1}

\AtBeginDocument{
\setlength{\abovedisplayskip}{0.5\abovedisplayskip}
\setlength{\abovedisplayshortskip}{0.5\abovedisplayshortskip}
\setlength{\belowdisplayskip}{0.5\belowdisplayskip}
\setlength{\belowdisplayshortskip}{0.5\belowdisplayshortskip}
}

\crefname{section}{Sec.}{Secs.}%
\crefname{appendix}{App.}{Apps.}%
\crefname{lemma}{Lem.}{Lemms.}
\crefname{theorem}{Thm.}{Thms.}
\crefname{corollary}{Cor.}{Cors.}
\crefname{equation}{Eq.}{Eqs.}
\crefname{figure}{Fig.}{Figs.}
\crefname{tabular}{Tab.}{Tabs.}

\newcommand{\X}{\ltlNext}
\newcommand{\F}{\ltlFinally}
\renewcommand{\G}{\ltlGlobally}
\tikzset{
	state/.style={
		rectangle,
		rounded corners,
		draw=black,
		minimum height=2em,
		minimum width=2em,
		inner sep=4pt,
		text centered,
	}
}

\usepackage[]{todonotes}

\begin{document}

\pagestyle{plain}
\maketitle
\setcounter{footnote}{0} 
\begin{abstract}
Synthesizing a reactive system from specifications given in linear temporal logic (LTL) is a classical problem, finding its applications in safety-critical systems design.
We present our tool \SemML, which won this year's LTL realizability tracks of \Syntcomp, after years of domination by \Strix.
While both tools are based on the automata-theoretic approach, ours relies heavily on (i) \emph{\textsc{Sem}antic labelling}, additional information of logical nature, coming from recent LTL-to-automata translations and decorating the resulting parity game, and (ii) \emph{\textsc{M}achine-\textsc{L}earning} approaches turning this information into a guidance oracle for on-the-fly exploration of the parity game (whence the name \SemML).
Our tool fills the missing gaps of previous suggestions to use such an oracle and provides an efficeint implementation with additional algorithmic improvements.
We evaluate \SemML{} both on the entire set of SYNTCOMP as well as a synthetic data set, compare it to \Strix, and analyze the advantages and limitations.
As \SemML{} solves more instances on SYNTCOMP and does so significantly faster on larger instances, this demonstrates for the first time that machine-learning-aided approaches can out-perform state-of-the-art tools in real LTL synthesis.
\end{abstract}
\section{Introduction}

\subsubsection{Synthesis} of finite systems from their logical specifications has been one of the central topics of theoretical computer science since the times of Church \cite{Church} and Büchi \cite{Buchi62}, being closely linked to developments of the automata theory \cite{DBLP:conf/stacs/Thomas95}.
Indeed, the logical formula would be translated to an automaton, in fact a game over this automaton played by the environment and system players, where the strategy of the latter corresponds to an implementation of the specified system.

Since Pnueli's suggestion to use Linear Temporal Logic (LTL) \cite{Pnueli77} for describing relevant properties of reactive systems, \emph{LTL synthesis} \cite{DBLP:conf/icalp/PnueliR89} has become an appealing alternative to manual implementation followed by LTL model checking.
Indeed, the tedious and error-prone implementation and debugging could be circumvented by automated construction of systems or their controllers, which are then correct \enquote{by construction}.
Nevertheless, the 2-EXPTIME-completeness of LTL synthesis, stemming from the doubly exponentially sized parity automata for the LTL formulae, has been challenging the practical applicability of the whole concept.
Fortunately, this has also led to numerous advances, such as identification of subclasses of properties for which the problem becomes easier, e.g.~\cite{DBLP:journals/tocl/AlurT04,DBLP:conf/vmcai/PitermanPS06,DBLP:conf/vstte/BansalGSLVZ22}, or methods avoiding the notoriously expensive step of determinizing the automata, e.g.~\cite{DBLP:conf/cav/KupfermanPV06,DBLP:journals/acta/TomitaUSHY17} or employing antichain-based methods, e.g.~\cite{DBLP:conf/cav/BohyBFJR12,DBLP:conf/tacas/CadilhacP23}.

The breakthrough of directly constructing deterministic automata orders of magnitudes \emph{smaller} \cite{DBLP:conf/cav/KretinskyE12,DBLP:journals/jacm/EsparzaKS20} has brought the classical automata-theoretic approach back on the stage, and indeed as the most efficient approach available.
This is witnessed by \Strix{} \cite{DBLP:conf/cav/MeyerSL18}, a synthesis tool based on the translations of Rabinizer/\Owl \cite{DBLP:conf/cav/KretinskyMSZ18,DBLP:conf/atva/KretinskyMS18} tools, winning the LTL tracks of the main synthesis competition \Syntcomp \cite{DBLP:journals/corr/abs-2206-00251}.

\subsubsection{Semantic Labelling and Previous Work.}
\begin{figure}[h]

\begin{tikzpicture}[x=4cm,y=1.5cm,font=\footnotesize,initial text=,outer sep=0.5mm]
	\tikzstyle{acc}=[double]
	\node[state,initial] (a) at (0,0) {$\neg a \vee \G\F(a\wedge\X b) ; a\wedge\X b$ };
	\node[state] (b) at (1.1,0) {$\G\F(a\wedge\X b); b$};
	\node[state] (c) at (2,0) {$\G\F(a\wedge\X b); \true$};
	
	\path[->] (a) edge node[above]{$a$} (b) (b) edge node[above]{$b$} (c)
	;
\end{tikzpicture}
\caption{An example of a part of an automaton with semantic labelling}
\label{fig:label}
\end{figure}
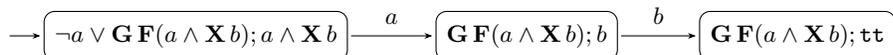

The dramatic improvements in the size came with an interesting side-effect.
In contrast to determinization of Safra \cite{DBLP:conf/focs/Safra88} and others \cite{DBLP:conf/lics/Piterman06,DBLP:conf/fossacs/Schewe09}, the new constructions are following the logical structure of the formula.
Consequently, the states of the generated automaton/game are labelled by this additional information.
It consists of the formula describing the property yet to be satisfied, i.e.\ monitoring the progress of satisfaction of the original formula, and formulae capturing progress of all its subformulae.
For example, an input formula $\neg a \vee \G\F(a\wedge\X b)$ labels the initial state of the automaton together with the sub-goal $a\wedge\X b$ to be satisfied infinitely often, see \cref{fig:label}.
After reading $a$, the successor state is labelled by the remaining goal $\G\F(a\wedge\X b)$ as well as the progressing sub-goal $b$ left to be satisfied.
Under $b$ the automaton then moves to the state with the first component remaining forever the same goal $\G\F(a\wedge\X b)$, but the second component, now being satisfied fully, signals that one repetition of the sub-goal has been successfully finished.

While this labelling was left unused for years, it clearly offers additional information.
Indeed, for instance, seeing $\neg a \vee \G\F(a\wedge\X b)$ as a goal, it seems easier to choose $\neg a$ in order to satisfy it than to take care of the infinitely repeating sub-goal.
Similarly, if progress is made in satisfying $a\wedge\X b$ by choosing an $a$, it seems wasteful not to follow with a $b$, although the overall goal of $\G\F(a\wedge\X b)$ remains unaffected either way.
Such a guidance can be used to explore the automaton/game on-the-fly and possibly finding a winning strategy before the whole state space is constructed.
In contrast, from the traditional perspective of solving games on graphs, one can either solve the whole game, or possibly explore it on-the-fly \enquote{blindly} since there is no observable difference between taking a transition, say, to the left or right.
Note that the guidance need not be reliable, the correctness is still guaranteed by solving (a part of) the game, hinting at possible use of machine learning.

In \cite{DBLP:conf/atva/KretinskyMM19}, two attempts have been made to explore the automaton/game in a profitable order using this additional information.
Firstly, the first component is subject to a na\"ive heuristic called \emph{trueness}, estimating the ease to satisfy a formula (by considering every formula as a Boolean combination, ignoring the temporal structure, and counting the percentage of satisfying assignments), and then transitions with higher trueness are explored first.
Secondly, reinforcement learning has been used with rewards being related to satisfying sub-goals in the second component.
While the former is also implemented in \Strix, both are ad-hoc heuristics with limitations.
In \cite{CAV23}, a machine-learning approach has been suggested, which learns from solving games for other formulae estimating which transition is more often \enquote{winning} than others.
This allows for superior precision, also dealing natively with more convoluted choices where the hand-written heuristic struggles.
However, only this oracle was implemented, not a (competitive) synthesis procedure.

\subsubsection{Our Contribution.}
In the present tool paper, we show how we incorporate this approach into the whole synthesis pipeline, closing the gaps explicitly left open.
Besides, we report on our tool efficiently implementing the approach and, even in a preliminary version, winning this year's edition of the realizability track\footnote{
	For realizability, the task is to determine whether a system satisfying the specification exists; its implementation, however, is only required in the synthesis track, where our tool did not participate.
	Competitively small representations of computed strategies require numerous (known) techniques, which are orthogonal to the advancements our tool is bringing into the area of machine-learning-aided solving of LTL games.
} of the \Syntcomp{} competition.
In more detail, our contribution is as follows:
\begin{itemize}[topsep=.5ex,itemsep=-.25ex,partopsep=.5ex,parsep=.5ex,left=.25em .. 1em]
	\item We implement a machine-learning heuristic guiding the on-the-fly exploration. 
	In contrast to \cite{CAV23} using SVM, we evaluate various models and choose the most adequate option. Besides, we adapt it to the state-of-the-art semantic labelling.
	\item We incorporate it into our synthesis pipeline, which improves over the approach of \Strix{} in several (traditionally algorithmic) aspects.
	\item We report on the performance of our tool \SemML{} (short for \textsc{Sem}antic-labelling-based Machine Learning) and analyze why it performs better than \Strix{}.
	It is worth noting that \SemML{} is faster on the \Syntcomp{} benchmark set while being trained only on synthetic data.
	On this synthetic data, it is even an order of magnitude faster.
\end{itemize}
Note that other lines of research that use LTL synthesis solvers as blackbox, e.g.\ LTL modulo theories synthesis \cite{RealizabilityModuloTheories} or portfolio solvers such as \Neurosynt{} \cite{DBLP:conf/tacas/CoslerHOS24}, directly profit from these improvements.

\subsubsection{Further Related Work.}

Besides \Strix{}, the closest to our work is, on the one hand, \textsc{Spot} (with \texttt{ltlsynt}) \cite{DBLP:journals/fmsd/RenkinSDP22}, following the same automata-theoretic approach, but constructing the whole automaton; and on the other hand, purely machine-learning approaches such as the deep-learning-based \cite{DBLP:conf/nips/SchmittHRF21,DBLP:conf/iclr/CoslerSHF23}, implemented in \Neurosynt{} \cite{DBLP:conf/tacas/CoslerHOS24}, which guesses circuits using ML, falling back to \Strix{} to achieve completeness.
Before the automata-theoretic approach, further winning approaches included bounded synthesis, e.g.~\cite{DBLP:conf/tacas/Ehlers11,DBLP:conf/cav/FaymonvilleFT17}, or even earlier safraless implementations \cite{DBLP:conf/fmcad/JobstmannB06}.
As mentioned, all of these are significantly out-performed by \Strix{} in \Syntcomp.

\section{Tool Description}

In this section, we provide an overview of our tool \SemML.
We formally state the problem it is solving, describe how to use the tool, and its high-level approach.

\subsection{Problem Description: LTL Synthesis and Realizability}
The problem of LTL reactive synthesis is defined as follows.
We are given an LTL \cite{Pnueli77} formula $\phi$ over a set of atomic propositions $\AP$ together with a partition of $\AP$ into \emph{environment} and \emph{system} propositions $\AP = \AP_{\smallenv} \union \AP_{\smallsys}$.
The environment and system generate an infinite word as follows.
In each step $i$, the environment chooses $e_i\subseteq  \AP_{\smallenv}$ and then\footnote{The convention that the environment chooses first and the system, observing the environment's choice, goes second is called \emph{Mealy} semantics.} the system chooses $s_i\subseteq \AP_{\smallsys}$, generating a sequence $e_1,s_1,e_2,s_2,\ldots$. 
The combined word over $\AP$ is then $e_1\cup s_1,e_2\cup s_2,\ldots$ and the system wins if this word satisfies $\phi$.
The central question is whether the system has a \emph{winning strategy}, i.e.\ a way to choose $s_i$ based on the current prefix so that the combined word always satisfies the given formula.\footnote{Formally, a function $f:(2^{\AP_{\smallenv}})^*\to 2^{\AP_{\smallsys}}$ so that $e_1f(e_1)e_2f(e_1e_2)\ldots\models \phi$.}
In that case, the instance is called \emph{realizable} and \emph{unrealizable} otherwise.
For example, the formula $\phi = \ltlGlobally (r \Leftrightarrow \ltlNext g)$ with $\AP_{\smallsys} = \{r\}$ and $\AP_{\smallenv} = \{g\}$ prescribes that whenever the environment sends a $r$equest, the system should in the next step $g$rant the request, and only then.
This formula is realizable, and a winning strategy is to remember whether the environment sent a request in the previous step.

Deciding whether a formula is realizable or not is called \emph{LTL realizability}.
In synthesis in the narrower sense, we want to output such a strategy for the winning player, i.e.\ a procedure that at every step outputs the next choice, typically in the form of a finite state machine, e.g.\ a \emph{Mealy machine} or an \emph{AIGER circuit} \cite{Syntcomp18}.
While our tool can output the strategy in a straightforward way, we refrain from discussing it further, as it neither the focus of the tool nor of our advancements.

\subsection{Functionality}
\paragraph{Inputs/Outputs}
\SemML{} accepts the standard format TLSF \cite{DBLP:journals/corr/Jacobs016} (used in SYNTCOMP), converted to LTL using \texttt{syfco}, as well as explicit input, i.e.\ an LTL formula together with a partition of the atomic propositions.
It supports both realizability and synthesis (with the strategy encoded as AIGER circuit).

\paragraph{Usage}
To streamline interaction, \SemML{} is invoked through a Python wrapper.
For TLSF input, use \texttt{main.py -{}-tlsf <path to tlsf file>}, and for explicit input \texttt{main.py -{}-ins=<ins> -{}-outs=<outs> -f=<formula>}, where \texttt{ins} and \texttt{outs} are the atomic propositions owned by environment and system, respectively.
The tool then solves the synthesis problem and outputs the witness strategy.
If desired, append \texttt{-{}-realizability} to only solve the realizability problem.
The tool then simply outputs \texttt{REALIZABLE} or \texttt{UNREALIZABLE}.

\begin{figure}[t]
	\centering
	\begin{tikzpicture}[
		component/.style={draw, rectangle,rounded corners=0.1cm, minimum height=1cm, minimum width=3cm,}]
		\node[] (input) at (0,1.25) {$\varphi$, $\AP_{\smallenv}\uplus\AP_{\smallsys}$};
		\node[] (output) at (4,1.25) {(UN)REALIZABLE};
		\node[component,align=center] (expl) at (0,0) {Frontier\\Exploration};
		\node[component,align=center] (ppgs) at (4,0) {Game\\Solver};
		\node[component] (bt) at (8,0) {Backtracking};
		
		\node [component,dashed,minimum height=2cm,minimum width=11.5cm] (box) at (4,-0.05) {};
		\node [anchor=south west] at (box.south west) {\SemML};
		
		\path [-latex]
		(input) edge (expl)
		(ppgs) edge node[pos=0.5,right] {\scriptsize if winner determined} (output)
		(expl) edge (ppgs)
		(ppgs) edge node[pos=0.5,above] {\scriptsize else}  (bt)
		(bt.south) edge[in=south,out=south,looseness=0.15] (expl.south)
		;
	\end{tikzpicture}
	\caption{
		High level architecture of \SemML.
	} \label{fig:architecture}
\end{figure}
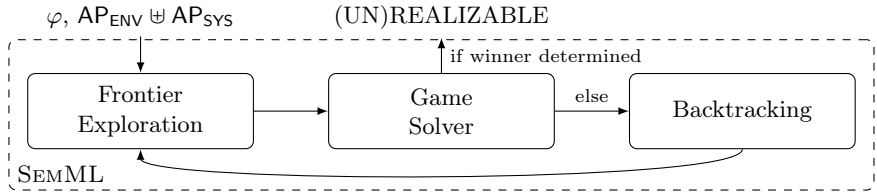

\subsection{High-Level Architecture} \label{sec:high_level}
In line with the automata-theoretic approach, \SemML{} employs \emph{on-the-fly} construction of the corresponding parity game.
In \SemML{}, this process comprises three main components, namely frontier exploration, partial game solving, and backtracking, also outlined in \cref{fig:architecture}.
In a nutshell, starting from an empty game, \SemML{} explores a \enquote{minimal} frontier. 
Here, we employ a sophisticated machine-learning (ML) guidance that, based on the semantic labelling, decides which parts of the game to explore first.
Then, our parity game solver tries to find a solution, interpreting unexplored states as losing for either player.
If the solver finds a solution, we are done.
If not, we need to explore more of the game.
We refer to a backtracking heuristic to identify candidate states and, starting from there, go back to frontier exploration.
In this setup, the main purpose of the new ML component is to tackle hard cases, where the automaton is too large to be constructed in its entirety, by trying to identify a small part where one of the players already wins.

\section{Advancements in Detail}
In this section we describe the major advancements of \SemML{}.
Recall that our technical goal is to employ ML to guide on-the-fly synthesis towards \enquote{easily winning} regions and thus improve scalability.
To describe our approach and contributions, we first discuss the state of the art.
Here, we consider the tool \Strix{}, which is currently the only implementation of this approach competitive on the standard \Syntcomp{} benchmarks, and
\cite{CAV23}, which provides the first ML-based approach to exploiting the semantic labelling.
In particular, we discuss their individual shortcomings and incompatibilities.
Then, we outline how we solve these issues, and describe our solution approach in detail.

\subsection{State of the Art and its Shortcomings}

\paragraph{\Strix{}}
alternates between exploring the parity game and trying to solve the explored part.
To decide which states to explore, \Strix{} uses a global double-ended priority queue (one end for each player) to track every state that can be explored further.
\Strix{} simply works on both ends of that queue while checking for solvability in fixed intervals until a winner is identified.
This comes with two major problems.
Firstly, the states are ordered by (a variant of) the rather na\"ive trueness \cite{DBLP:conf/atva/KretinskyMM19} of formulae, which roughly corresponds to the percentage of \emph{propositionally} satisfying assignments, completely disregarding the temporal structure.
This is particularly problematic for formulas which comprise lots of temporal behaviour (such as $\ltlNext \ltlNext \phi$).
Secondly, the exploration is not \enquote{demand-driven}, meaning it can fail
because one successor of an important state has not been explored (and thus is considered losing), but during the subsequent exploration phase, that successor is not touched either because other states have more extreme scores (or many states have the same score).

\paragraph{The approach of \cite{CAV23}}
uses the semantic labelling to predict winning choices locally through a simple learning approach with hand-crafted features.
These are then used as the starting point for a parity game solver, ensuring that potential imprecisions of the ML model are detected and fixed.
However, (i)~they do not provide an implementation competitive w.r.t.\ the actual runtime, and (ii)~adapting their approach to \Strix{} (or any other tool) is difficult for a variety of reasons.
Firstly, \Strix{} is fundamentally designed to work with a \enquote{global} ranking, i.e.\ picking states to explore from one priority queue, while \cite{CAV23} gives \enquote{local} recommendations, specific to a concrete state.
Note that combining the two approaches by first picking a state similarly to \Strix{} and then following \cite{CAV23} in that state does not solve \Strix{}'s lack of demand-driven guidance.
Secondly, \cite{CAV23} employs complex features and evaluating them is too time consuming.
The timing constraints on evaluating the guidance heuristic are quite strict, since we need to be able to give hundreds of recommendations per second to remain competitive.
Finally, \cite{CAV23} uses an outdated automaton construction (via LDBA \cite{Sickert_16_LDBA,DBLP:journals/sttt/EsparzaKRS22}), while \Strix{} uses a newer, practically more efficient variant.
This is particularly relevant, as the automaton construction usually is the biggest bottleneck of LTL synthesis. 

\subsubsection{Summary.}
In order to reap the benefits of ML for on-the-fly synthesis, we thus first have to design a novel exploration approach (and adapt all subsequent machinery) capable of processing \emph{local advice} in the spirit of \cite{CAV23}.
While this is a pre-requisite for using ML guidance, it is also interesting in its own right, as it allows for a more targeted, deep exploration instead of exploring multiple equally promising directions simultaneously.
Then, at the same time, our guidance must be much more efficient than \cite{CAV23}, so that it does not add too much overhead, which would negate any performance gained by giving good recommendations.
Finally, it also needs to be designed for a modern automaton construction.

We proceed to explain how we tackled these problems.
We introduce our locally guided approach to on-the-fly LTL synthesis, describe how we use machine learning to guide the exploration, and finally outline general engineering improvements.

\subsection{Locally Guided Exploration} \label{sec:tool:explore}
Recall that the overall approach is to alternate between exploring parts of the parity game and checking whether there exists a winning strategy in the current part of the game already, as depicted in Fig.~\ref{fig:architecture}.
For the exploration, our aim is to follow \enquote{good} choices for one player that work well against all options of its opponent.
As such, we run the exploration for both \enquote{perspectives} separately, and regularly switch between them (further motivation and details in \cref{sec:eng}).
In the following, we take the perspective of the system (aiming to prove realizability); the dual part for the environment is analogous.

\begin{figure}[t]
	\centering
	\vspace*{-2em}
	\begin{tikzpicture}[component/.style={draw, rectangle,rounded corners=0.1cm, minimum height=1cm, minimum width=2.5cm}]
		
		\node (initState) at (2,3) {initial state};
		\node[align=center] (BTSugg) at (6,3) {Backtracking\\suggestion};
		\node[align=center] (solver) at (10,3) {Game Solver};
		
		\node[cylinder,draw,minimum width=1cm,minimum height=3cm] (queue) at (6,1.5) {Frontier Queue};
		\node[component,align=center] (aut) at (3,0.3) {Automaton \\ Construction};
		\node[component,align=center] (game) at (6,0.3) {Game \\ Construction};
		\node[component,align=center] (eh) at (9,0.3) {Exploration \\ Heuristic};
		
		\draw[-latex] (queue) -- (1,1) -- (1,0.3)  node[below] {pop} -- (aut);
		\draw[-latex] (eh) -- (11,0.3) node[below] {add}  -- (11,1) -- (queue);
		
		\path[-latex]
		(initState.south) edge[in=north,out=south,looseness=0.5] ([xshift=-0.5cm]queue.north)
		(BTSugg.south) edge[in=north,out=south] (queue.north)
		([xshift=0.5cm]queue.north) edge[in=west,out=north,looseness=0.7] node[below right,pos=0.5] {\scriptsize if game closed} (solver.west)

		(aut) edge (game)
		(game) edge (eh)

		;
		
		\node[component,dashed,minimum height=2.7cm,minimum width=11.5cm] (box) at (6,1) {};
		\node [anchor=north west] at (box.north west) {\textit{Frontier-Exploration}};
	\end{tikzpicture}\\[0.1cm]
	\begin{tikzpicture}[component/.style={draw, rectangle,rounded corners=0.1cm, minimum height=1cm, minimum width=2.5cm},autState/.style={draw,circle,font=\tiny, inner sep=1pt},
		sysState/.style={draw,rectangle,font=\tiny,inner sep=3pt}]
		\node[autState] (q-10) at (1,-1.7) {$q_0$};
		
		\node[autState] (q00) at (2,-1.7) {$q_0$};
		\node[autState] (q01) at (4,-1.2) {$q_1$};
		\node[autState] (q02) at (4,-2.2) {$q_2$};
		\path [-latex']
		(q00) edge (q01)
		(q00) edge (q02)
		;
		
		\node[autState] (q10) at (5,-1.7) {$q_0$};
		\node[autState] (q11) at (7,-1.2) {$q_1$};
		\node[autState] (q12) at (7,-2.2) {$q_2$};
		\node[sysState] (s11) at (6,-1.2) {$s_1$};
		\node[sysState] (s12) at (6,-2.2) {$s_2$};
		\path [-latex']
		(q10) edge (s11)
		(q10) edge (s12)
		(s11) edge (q11)
		(s11) edge (q12)
		(s12) edge (q11)
		(s12) edge (q12)
		;
		
		\node[autState] (q20) at (8,-1.7) {$q_0$};
		\node[autState,dashed] (q21) at (10,-1.2) {$q_1$};
		\node[autState] (q22) at (10,-2.2) {$q_2$};
		\node[sysState] (s21) at (9,-1.2) {$s_1$};
		\node[sysState] (s22) at (9,-2.2) {$s_2$};
		\path [-latex']
		(q20) edge (s21)
		(q20) edge (s22)
		(s21) edge[dashed] (q21)
		(s21) edge[thick] (q22)
		(s22) edge[dashed] (q21)
		(s22) edge[thick] node[below] {\scriptsize EH}  (q22)
		;
		
		\node[] at (11,-1.7) {add($q_2$)};
		
		\node[component,dashed,minimum height=1.9cm,minimum width=11.5cm] (box) at (6,-1.75) {};
		\node [anchor=south west] at (box.south west) {\textit{Example}};
		
	\end{tikzpicture}
	\caption{
		Illustration of the exploration process together with an example for each step.
	} \label{fig:frontier_structure}
\end{figure}
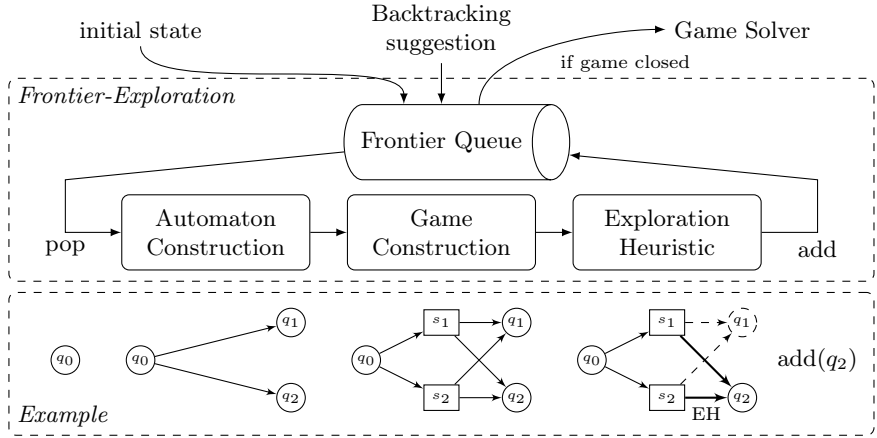

As hinted in \cref{sec:high_level}, our exploration approach comprises two parts, namely frontier exploration and backtracking.
The components of frontier exploration and its interplay with both backtracking and the game solver is depicted in \cref{fig:frontier_structure}.
During frontier exploration, the core idea is to only explore a necessary minimum so that a strategy of the system can be at all properly defined.
In particular, we want to reach a point where every known system state has \emph{at least one} of its successors explored and every known environment state has \emph{all} of its successors explored.
If that is the case, we call the (partial) parity game \emph{closed}.
Clearly, to obtain a closed game, we repeatedly need to explore states.
To this end, we maintain a queue of automaton states (the current frontier) which we still need to explore.
After taking a state from the queue, we compute the immediate automaton successors, using an adapted implementation of \Owl, and split this automaton transition (under a subset of $\AP$) into the two moves of the players under a subset of $\AP_{\smallenv}$ and $\AP_{\smallsys}$, respectively.
While we hardly have a choice for environment states (we need to explore all successors in the game), in system states we can select which successor to explore.
Thus, in these states we ask our exploration heuristic for advice, and add all newly reached states to the queue.
For this local guidance, we employ the new ML-based approach, which we later explain in depth.
For an example, see \cref{fig:frontier_structure} (bottom).
From left to right, we obtain a state $q_0$ from the queue and construct its successors $q_1$ and $q_2$ in the automaton.
Splitting it into the game introduces the two system states $s_1$ and $s_2$.
In both states, the exploration heuristic recommends going towards $q_2$, and we only add that state to the frontier queue.

Overall, we repeat this process until the current game is closed, and then attempt to solve it.
If we cannot determine a winner, then in at least one of the system states a \enquote{wrong} successor was chosen (the system cannot win the current partial game).
Thus, subsequently, we ask the backtracking oracle which states might have been \enquote{wrong}.
Concretely, we heuristically choose a subset of all non-fully explored states with the highest trueness.
For each of these, we explore their next best successors according to our heuristics.
Now, the game might not be closed, and thus we switch back to frontier exploration.
We repeat this process until a winner is found (which always happens, as eventually the entire game is explored).

\subsection{Exploration Guidance Through Machine Learning} \label{sec:tool:ml}
In this section, we describe our ML approach used to guide the frontier exploration.
Recall that for a given state the exploration heuristic is supposed to give a ranking preferring \enquote{good} edges which lead to a winning strategy:
Initially, we follow the highest ranked choice, then, when backtracking in this state, the second one, and so on.

Thus, we would like this heuristic to prefer edges that can be part of a winning strategy.
Additionally, as we also want to obtain small games, among the possibly winning edges we would like to prefer edges leading to smaller strategies (hence exploring a smaller part of the game).
Note that one can also employ handcrafted heuristics instead, e.g.\ the score computed by \Strix{} (which we also implement and evaluate).

Similar to \cite{CAV23}, we employ a supervised learning approach.
As usual for ML, we start by describing the dataset used to train our models.
We then discuss the overall architecture of the model(s), how we obtain the ground truth, and how we extract features.
Finally, we discuss the training method.
Of course, our approach is not the only possible way to tackle this problem.
Yet, while designing it, we discovered several subtle pitfalls and tried alternative approaches which proved to be suboptimal.
We provide further details within this section.

\subsubsection{Data}
Our explicit aim is to exploit structure in real-world formulae.
Here, existing datasets such as the \Syntcomp{} set seem a natural choice.
However, we want to evaluate our approach on the entire \Syntcomp{} set (in order to faithfully replicate the \Syntcomp{} evaluation).
Thus, \enquote{showing} any part of it during learning could introduce an unfair advantage.
This leaves us with hardly any realistic data sets.

This problem has already been observed by \cite{DBLP:conf/nips/SchmittHRF21}.
As a solution, they note that in practice, specifications often follow specific \emph{patterns} and combinations thereof \cite{specificationPatterns}.
Thus, randomly combining such patterns should yield numerous formulae that resemble some structure one might expect in practice.
To this end, \cite{DBLP:conf/nips/SchmittHRF21} identified over 150 \emph{assumptions} and over 1500 \emph{guarantees} which intuitively limit the behaviour of the environment and system respectively.
From this set of \enquote{building blocks}, they generate formulae by sampling assumptions and guarantees and assemble them in the form of \enquote{conjunction of assumptions implies conjunction of guarantees}, which adds some comprehensible structure.
Formulae of this kind can be interpreted as \enquote{if the environment adheres to one behaviour profile, the system should adhere to another}.

We extend this idea a bit further by sampling several options for system and environment, which diversifies the formulae while maintaining comprehensibility.
In particular, we sample multiple sets of assumptions and guarantees and assemble a formula in the form of \enquote{DNF of assumptions implies DNF of guarantees}.
Intuitively, these formulae mean \enquote{If the environment follows one of these behaviour profiles, the system should adhere to one of their behaviour profiles}.
In particular, this introduces options for the system to which behaviour profile it should adhere to, which in turn might depend on the profile the environment chooses.

We filter our generated data into two groups depending on the size of the corresponding automaton.
The training and validation group consists of 1000 formulae where the automaton has at most 500 states.
We introduce this limitation to keep the ground truth and feature computation feasible (described later).
While 1000 formulae may seem like a small data set, note that we learn from the local decisions in each state of the 1000 associated parity games, which give several million data points in total.
For evaluation, we also identified 200 formulae of which the automaton size is not known except that it is larger than 20{,}000 states.

While this yields a decent data foundation for our venture, the synthetic data definitely is quite different from \Syntcomp.
Thus, for practical purposes, one should consider including \Syntcomp{} and other data during learning.
Since this is orthogonal to the evaluation in this paper (including any part of \Syntcomp{} in our training data would introduce unwanted biases), we deliberately do not include this.

\subsubsection{Model Architecture}
Similar to \cite{CAV23}, we rank outgoing edges through all pairwise comparisons.
Formally, we employ a \emph{pair classifier} $p: E\times E \to \Reals$ where the sign denotes whether the first or second edge is preferred and the magnitude denotes the confidence in that prediction.
In a state, every pair of edges is compared and each edge is ranked according to the sum of confidences in its favour.
However, since this scales quadratically in the number of outgoing edges, we approximate the above for states with more than 16 edges.
For them, a number of \emph{pivot edges} are chosen that every other edge is compared to, in order to obtain a \enquote{first guess} at a ranking.
From that guess, the best 8 are selected to enter the second round which now is a full round of comparisons.
The final ranking comprises the ranking among the top 8 followed by the other edges according to the first ranking.

Moreover, similar to \cite{CAV23}, we pre-classify states into groups with conceptual differences and train a separate model for each group.
Intuitively, we distinguish (i)~whether a state is owned by the system or the environment, and (ii)~whether the long-term goals are trivially structured, e.g.\ a single liveness condition, which simplifies some decisions; leading to 4 models in total.
We discuss the concrete implementation of the pair classifier in the \enquote{Training} section below.

\begin{figure}[t]
	\centering
	\vspace*{-2em}
	\begin{tikzpicture}
		[sysState/.style={draw,rectangle,font=\small,inner sep=3pt, minimum width=1cm,minimum height=0.5cm}]
		\node[sysState, initial left] (1) at (0,0.6) {$u$};
		\node[sysState] (2) at (3,0.6) {$v_1$};
		\node[sysState] (3) at (3,-0.6) {$v_2$};
		\node[sysState] (4) at (1.1,-0.3) {$w$};
		\node[sysState] (T) at (6,0) {\text{goal}};
		
		\path[->]
		(1) edge (2)
		(1) edge[bend right=12] (4)
		(4) edge[bend right=12] (3)
		(2) edge[bend right=20] (3)
		(3) edge[bend right=20] (2)
		(2) edge[bend left=8] (T)
		(3) edge[bend right=8] (T)
		(T) edge[loop above] (T)
		;
	\end{tikzpicture}
	\caption{
		A simple game to illustrate two challenges for the ground truth.
		For simplicity, all states are controlled by one player.
		Clearly, all states are winning.
	}\label{fig:SI_bad}
\end{figure}
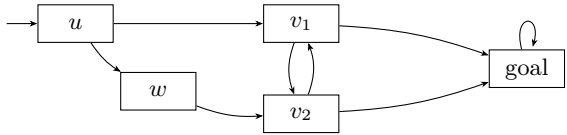

\subsubsection{Ground Truth}
For the supervised learning of our models, we need meaningful labels that denote the quality of an edge so that we can determine the better one of any given pair.
But which edges are \enquote{good}?
At first, this may seem obvious -- simply take all edges which are part of a winning strategy.
This however is problematic for multiple reasons, as already outlined in \cite{CAV23}.

First, parity games do not allow for \emph{maximally permissive} strategies.
This means that there simply is no one \enquote{local truth}; whether an edge is good or bad may depend on decisions in other states, as we exemplify in \cref{fig:SI_bad}.
While the edges leading to the goal from $v_1$ and $v_2$ are always winning choices, edge $\overline{v_1v_2}$ is only winning if the goal-edge is chosen in $v_2$ (and vice-versa for $\overline{v_2v_1}$). 
Consequently, different solution approaches may yield different sets of winning edges, and just considering one of them would bias the model to behave alike to \emph{that concrete solution method} and not to \enquote{understand} semantically labelled parity games in general.
Relating to the previous example, even though $\overline{v_1v_2}$ and $\overline{v_2v_1}$ are symmetric, a solver may only mark one of them as winning, but never both.
As such, using the output of one solver would actively try to make our model believe that one of the two is better and imitate that solver's bias.

Secondly, even if multiple edges are indeed winning, this does not inform us about the \enquote{complexity} required to win after playing one of them.
For example, consider two edges where one leads to a trivially winning sink within two steps and the other leads to a large and complicated, but ultimately also winning region.
Qualitatively, both edges are equivalent, but we prefer the former as it yields smaller solutions and requires fewer correct decisions in the future.
We also provide an illustration in \cref{fig:SI_bad}.
There, both choices in $u$ are winning, however we prefer moving to $v_1$ over moving to $w$, as we can win \enquote{faster}.

Thus, in order to determine the quality of any given edge of the game, we analyze the game tree after playing said edge.
Constructing the entire game tree and applying min-max is practically infeasible already for rather small instances.
Therefore, we apply an improved version of the decayed Monte Carlo tree search suggested in \cite{CAV23}.
In particular, we only deeply expand the tree for critical paths (the ones where either player fancies their chances) and thus can identify longer shortest winning paths.
Conveniently, we thereby no longer require the \enquote{optimal stalling} strategy that \cite{CAV23} uses for the opponent, as the opponent prefers longer losing paths over shorter ones by default due to the decay.
In the end, we effectively get a score between $-1$ and $1$ for each edge in the game which indicates the \enquote{quality} of this edge.
Intuitively, an edge directly leading to \texttt{tt} gets a $1$.
An edge leading to a region where the system can win but may require a lot of steps to do so yields a small, positive value, while edges after which the environment can quickly force a losing cycle yield a score close to $-1$.

\subsubsection{Feature Extraction}
The feature extraction transforms a transition in the game into a vector of numbers so that it can be processed by an ML model.
The features are based on all the information that is available at the time of deciding which edge to explore further.
In particular, this includes the semantic labelling of the transition's source and target, the colour/priority, and also labelling associated to its sibling transitions.

We deliberately aimed at manually designing a (large) set of features derived from the semantics and then prune it via feature selection.
While automatic feature extraction is a powerful tool, in our use case the feature extraction needs to be extremely efficient, since we need to call it hundreds of times per second to remain even remotely viable.
(Recall that we need to extract the features for every edge in the game and already the games obtained from reasonably simple formulae can easily reach thousands of states and significantly more edges.)

\paragraph{State Features}
We first introduce twelve \enquote{formula features}, which transform a single LTL formula into a number.
Intuitively, they can be thought of as proxies for higher level concepts.
These concepts include \emph{formula-complexity} (syntactic properties such as height and size of the syntax tree), \emph{formula-sat-difficulty} (how \enquote{difficult} is it to satisfy the formula, capturing variants of \emph{trueness} \cite{DBLP:conf/atva/KretinskyMM19}), or \emph{formula-controllability} (how much influence does a player have on the truth value of the formula with only their variables).
Formula features are then aggregated for a \emph{state} (which comprises several formulae) in one of two ways.
Either, we select a single formula of the labelling and yield the value of the base feature on the selected formula as the state's overall value (e.g.\ selecting the formula that maximizes the value of another base feature).
Or, we apply the base feature to all formulae of the semantic labelling and aggregate the results in several ways, exploiting the non-trivial structure of the state labelling.
Intuitively, this captures the respective concept (e.g.\ controllability) over the entire state.

\paragraph{Edge Features}
Edge features are obtained from state features by either taking the state feature of the edge's successor or the change of the feature along the edge, i.e.\ the difference of the feature in the successor and predecessor.
Further, we can compare that value against the value of all other edges of the same state by normalizing the feature to the $[0,1]$-interval, so that a normalized value of $1$ denotes that it is the highest among its sibling edges.
This may help learning relative comparisons, but loses all information on the absolute value of the feature.
As confirmed in our final models, a mixture of normalized and non-normalized features seems to be desirable.
Aside from features derived from states, we also consider features based on the edge priority as suggested by \cite{CAV23}.
However, we include the parity information in an \enquote{ML-friendly} way, for example by mapping it to a linear scale.

In total, we obtain well above 150k different features for edges.
These include, for example, the change in the syntax tree height of the most controllable formula or the aggregated trueness, normalized across all successors.
For more details on the features we refer to \iftoggle{arxiv}{\Cref{app:features}}{the appendix of \cite{arxivVersion}}.

\subsubsection{Training}\label{sec:Training}
Applying ground truth and feature methods to the generated formulae yields a dataset for supervised learning with way over a million samples for every state class.
We bootstrap these down to roughly 100k samples per state class in order to make the training take reasonable amounts of time.
The following procedure was done for every state class individually.

\paragraph{Feature Elimination}
As using our entire set of thousands of features is absolutely impractical in several regards, especially with our performance constraints in mind, we perform multiple stages of feature elimination.
First, we randomly select between 50 an 100 features per major category and add some hand-picked features.
This leaves us at an algorithmically manageable, yet practically infeasible amount of about 700 features.
For further reductions, we perform a variant of recursive feature elimination, adjusted to our pairwise setting (see \iftoggle{arxiv}{\Cref{app:model_experiments}}{\cite{arxivVersion}} for details).
As different features might be more or less important for different model types and state classes, we ran a separate feature elimination for each of these.
For details on what kind of features remained after the elimination, we refer to \iftoggle{arxiv}{\Cref{app:features}}{\cite{arxivVersion}}.

\paragraph{Models}
As for model types, we evaluated (kernel-)SVMs, neural networks, random forests, and gradient boosted trees.
The input for each model is the concatenation of the two feature vectors of the respective edges that we want to compare pairwise.
For tree models, we additionally concatenated the pointwise differences of the features in order to allow them to compare the same feature of both edges in one decision node.
For every model type, we performed several smaller runs to obtain suitable hyper-parameters for the large scale feature elimination.

Ultimately, gradient boosted trees proved to be the best choice for implementing our pair classifier in all four state classes.
Together with random forests, they clearly outperformed the non-tree methods like SVM or NN.
However, in contrast to random forests, they required less features (3-10, depending on state class) to do so.
Further details on this experiment can be found in \iftoggle{arxiv}{\Cref{app:model_experiments}}{\cite{arxivVersion}}.

\subsection{Engineering}\label{sec:eng}
To conclude, we provide details on our implementation and engineering improvements.
First and foremost, as a major practical improvement, \SemML{} is implemented in pure Java (built on top of \Owl).
In contrast, \Strix{} is developed as a hybrid between Rust and Java, with native compilation of Java through GraalVM, and a complex interplay between the two code bases.
This adds, among others, complexity due to working with two languages, subtle performance overheads when crossing language boundaries, and setup difficulties due to requiring a rather particular set of tools.
As such, \SemML{} is significantly easier to use, maintain, and extend.
For easy incorporation of third party tools such as \texttt{syfco}, we also include a Python wrapper.
For learning, we use the Python library \textsc{sklearn} \cite{scikit-learn}, and store our models in the established PMML format.

Aside from structural improvements and pure-Java implementation, we also added several engineering changes compared to the approaches of \Strix{} and \cite{CAV23}, of which we list a few notable ones.

\begin{description}
	\item[State Merging]
	In parity games, states are fully determined by their set of edges, i.e.\ if two system states transition to the same successor for every system assignment while emitting the same priority, they are equivalent and can be merged.
	This equivalence check is very efficient due to our internal representation of states.
	Interestingly, by applying this reduction we observed a decrease of game sizes by up to two orders of magnitude.

	\item[Dual Perspectives]
	As mentioned in \cref{sec:tool:explore}, our approach alternates between the perspectives of both players.
	While this is not required for correctness, it helps in practice, as, for example, we can quickly find a small winning region for the environment even if the system player explores in a different direction.

	\item[Exploration Scheduling]
	Usually, we switch perspectives whenever the game is closed and the solver is consulted.
	However, when following a \enquote{bad} edge leads the algorithm off the track, we might spend a lot of time trying to close the game.
	Instead of insisting on continuing this process to the end, we also switch to the other perspective after too many states have been explored without closing the game.

	\item[Result Sharing]
	We re-use information discovered from one \enquote{perspective} for the other part, where appropriate, for example already constructed parts of the automaton.
	Moreover, if one side finds a set of states to be winning for them, the exploration of the other side directly treats them as losing.

	\item[Caching]
	We trade memory for time by caching all computed features.

	\item[BDD]
	We implemented complement edges and several further engineering improvements in the underlying pure-Java BDD library \textsc{JBDD} \cite{jbdd}.
\end{description}

\section{Experimental Evaluation}\label{sec:experiments}

In this section, we evaluate the two central research questions of interest, namely:
\begin{description}
	\item[RQ1] Can \SemML{} solve LTL realizability more efficiently than state-of-the-art tools, in particular \Strix{}?
	\item[RQ2] How much of the improvements are caused by algorithmic and engineering changes and how much by employing ML-guided exploration?
\end{description}
We first introduce considered tools, metrics of interest, and our benchmark sets.
Then, we present our results and discuss each research question separately.

\paragraph{Tools}
We consider our tool \SemML{} and the state-of-the-art tool \Strix{}\footnote{With its best-performing configuration \texttt{-{}-exploration=minmax -{}-lookahead=0}.}.
To further distinguish algorithmic and engineering improvements from those due to the ML-based exploration heuristic, we also consider \SemMLa{} (\enquote{\SemML{} without ML}), which uses the exploration score of \Strix{} as guidance instead.

Note that both \SemML{} and \Strix{} internally construct a strategy even when \enquote{only} solving LTL realizability.
The problem of exporting the (already constructed) strategy into a particular format, e.g.\ AIGER circuits, is completely orthogonal.
Thus, we explicitly focus on the time to find the solution (i.e.\ let the tools run in their \enquote{realizability} configuration).

\begin{remark}
	When only focusing on the number of solved instances, the portfolio solver \Neurosynt{} \cite{DBLP:conf/tacas/CoslerHOS24} is a more powerful tool than \Strix{}.
	This is to be expected, as it runs multiple approaches (including \Strix{}) in parallel.
	However, to clearly compare our specific approach to the state-of-the-art, we deem a direct comparison of \SemML{} to \Strix{} (the state-of-the-art of \enquote{single-approach}-solvers) more relevant.
	A more detailed discussion and comparison of \SemML{} with \Neurosynt{} can be found in
	\iftoggle{arxiv}{\Cref{app:neurosynt}}{the appendix of \cite{arxivVersion}}.
\end{remark}

\paragraph{Metrics} 
Primarily, we are interested in which tool can solve more benchmarks within a given time constraint (the main metric used for \Syntcomp).
Additionally, for the instances solved by multiple tools, we are interested in which tool solves them faster.
For that matter, we compute the ratios of (wallclock) time for all samples that both tools were able to solve and aggregate them by computing the geometric mean.\footnote{As is customary, the geometric mean is preferable for runtime ratios:
For example, for 0.5x and 2x speed-ups it yields 1x instead of 1.25x with an arithmetic mean.}
We exclude simple instances to account for constant time overheads caused by, e.g., JVM startup and loading of ML models.
We treat an instance as simple if both compared tools solve them faster than a given threshold (usually 5s; later we also consider larger values to focus on the most complicated instances).

\paragraph{Benchmarks}
We consider two classes of benchmarks.
First, our evaluation set as described in \cref{sec:tool:ml}, called Synthetic.
Here, we expect the ML-based guidance to shine, as the inputs are of a similar structure as the training data (just much larger).
Additionally, we consider the entire set of SYNTCOMP 2024.

We highlight some peculiarities of the SYNTCOMP data set.
First, as we observe in our experiments, the vast majority of instances are trivial, i.e.\ solved within a few seconds.
As such, a small constant time overhead has a large (relative) impact.
Second, SYNTCOMP mainly comprises parametrized families of formulae, where incrementing the parameter often more than doubles the size of the state space. 
As such, for many families a lot of their instances are simple, very few are interesting-but-solvable, and then many more are completely out of reach.
This results in a rather small set of benchmarks where a notable difference can be expected, and solving one more sample of a family already marks significant improvement. 
Finally, there are subtle biases and asymmetries that may limit the possible improvement of exploration guidance.
On the one hand, several of the realizable families are arbiters (or variants thereof), where, by design, (nearly) the entire state space needs to be explored.
In particular, any attempt of guidance is useless and any effort spent on it costs overall performance.
On the other hand, many unrealizable families are constructed by taking a realizable family and introducing a contradiction at a parametrized depth of the execution.
This class may be more suitable for employing targeted guidance, as one only needs to find that single contradiction in the state space to prove unrealizability.
However, these unrealizable families tend to be dominant in numbers while not providing much diversity.
Thus, if a heuristic by chance adapts well or badly to a single family or a particular kind of contradiction, this effect alone can dominate the overall evaluation.
As such, the results on \Syntcomp{}, should be interpreted carefully, especially when comparing guidance heuristics.

\paragraph{Experimental Setup}
Our experiments were conducted on an AMD Epyc 7443 24-Core CPU and 188GB of RAM.
Each invocation was limited to 30 minutes and 60GB memory, mimicking SYNTCOMP conditions.

\begin{table}[t]
	\caption{
		Comparison of \SemML{} to \Strix{} on \Syntcomp{} and synthetic data.
		We show how many instances are solved by both, by only one, or none of the tools.
		Note that \trivialStrixSemmlSyntcompTen{} instances of \Syntcomp{} are solved by both tools in under 10 seconds.
	} \label{tbl:results}
	\centering
	\setlength{\tabcolsep}{4pt}
	\renewcommand{\arraystretch}{1.1}
	\begin{minipage}{0.49\textwidth}
		\centering
		\begin{tabular}{lccc}
			        \multicolumn{2}{c}{\multirow{2}{*}{\Syntcomp}}          &                    \multicolumn{2}{c}{\SemML}                    \\
			                                                     &          &            solved             &             unsolved             \\
			\midrule
			\multirow{2}{*}{\rotatebox[origin=c]{90}{\Strix}} &  solved  & \StrixSemmlSyntcomptfZeroAllM &  \StrixSemmlSyntcomptfZeroAllK   \\
			                                                     & unsolved & \StrixSemmlSyntcomptfZeroAllL & \unknownStrixSemmlSyntcomptfZero
		\end{tabular}
	\end{minipage}
	\begin{minipage}{0.49\textwidth}
		\centering
		\begin{tabular}{lccc}
			       \multicolumn{2}{c}{\multirow{2}{*}{Synthetic}}        &                   \multicolumn{2}{c}{\SemML}                   \\
			                                                  &          &            solved            &            unsolved             \\
			\midrule
			\multirow{2}{*}{\rotatebox[origin=c]{90}{\Strix}} &  solved  & \StrixSemmlSyntheticZeroAllM &  \StrixSemmlSyntheticZeroAllK   \\
			                                                  & unsolved & \StrixSemmlSyntheticZeroAllL & \unknownStrixSemmlSyntheticZero
		\end{tabular}
	\end{minipage}
\end{table}

\begin{table}[t]
	\caption{
		Average runtime ratios between \Strix{} and \SemML{} for all instances where at least one tool required more than $n$ seconds (where $n$ is in the top row) and both tools found a solution.
		We also give the number of instances that satisfy these two criteria.
		A ratio $>1$ indicates that \SemML{} is faster on average.
	} \label{tbl:speed}
	\centering
	\setlength{\tabcolsep}{5pt}
	\renewcommand{\arraystretch}{1.1}
	\begin{minipage}{0.49\textwidth}
	\centering
	\begin{tabular}{ccccc}
		\Syntcomp &               0               &               5               &               30                &                  300                  \\
		\midrule
		  ratio   & \StrixSemmlSyntcomptfZeroAllT & \StrixSemmlSyntcomptfFiveAllT & \StrixSemmlSyntcomptfThirtyAllT & \StrixSemmlSyntcomptfThreehundredAllT \\
		  count   & \StrixSemmlSyntcomptfZeroAllM & \StrixSemmlSyntcomptfFiveAllM & \StrixSemmlSyntcomptfThirtyAllM & \StrixSemmlSyntcomptfThreehundredAllM
	\end{tabular}
	\end{minipage}
	\begin{minipage}{0.49\textwidth}
	\centering
	\begin{tabular}{ccccc}
		Synthetic &              0               &              5               &               30               &                 300                  \\
		\midrule
		  ratio   & \StrixSemmlSyntheticZeroAllT & \StrixSemmlSyntheticFiveAllT & \StrixSemmlSyntheticThirtyAllT & \StrixSemmlSyntheticThreehundredAllT \\
		  count   & \StrixSemmlSyntheticZeroAllM & \StrixSemmlSyntheticFiveAllM & \StrixSemmlSyntheticThirtyAllM & \StrixSemmlSyntheticThreehundredAllM
	\end{tabular}
\end{minipage}
\end{table}

\subsection{RQ1: Comparing \SemML{} to \Strix{}}
\begin{figure}[t]
	\centering%
	\pgfplotsset{
		table/col sep=comma,
		plot/.style={
			width=\textwidth,height=\textwidth,
			x label style={anchor=north,inner sep=0pt},
			y label style={anchor=south,inner sep=0pt},
			xtick={2,60,1800}, xticklabels={2,60,1800}, 
			ytick={2,60,1800}, yticklabels={2,60,1800},
			xmin=1,ymin=1,xmax=\legendmax,ymax=\legendmax,
			axis x line*=bottom,
			axis y line*=left
		}
	}%
	\begin{minipage}{0.5\textwidth}
		\centering
		\renewcommand{\plotmarksize}{2pt}
		\begin{tikzpicture}
			\begin{axis}[plot,xmode=log,ymode=log,xlabel=\Strix,ylabel=\SemML]
				\axislines
				
				\syntheticplot table [x=time_Strix, y=time_Semml] {data/time/Synthetic_Strix_Semml_all.csv};
				\syntunrealplot table [x=time_Strix, y=time_Semml] {data/time/SyntcompTF_Strix_Semml_real.csv};
				\syntunrealplot table [x=time_Strix, y=time_Semml] {data/time/SyntcompTF_Strix_Semml_unreal.csv};
			\end{axis}
		\end{tikzpicture}

		\vspace{0.25cm}\hspace*{1cm}
		\begin{tikzpicture}[baseline]
			\node[draw,rectangle,inner sep=4pt] {~\markincaption{Dark2-B}{star}~\Syntcomp{} \quad \markincaption{Dark2-C}{asterisk}~Synthetic~};
		\end{tikzpicture}
	\end{minipage}\hspace{0.5cm}
	\begin{minipage}{0.32\textwidth}
		\begin{tikzpicture}
			\begin{axis}[plot,xmode=log,ymode=log,xlabel=\Strix,ylabel=\SemMLa]
				\axislines
				
				\syntheticplot table [x=time_Strix, y=time_SemmlA] {data/time/Synthetic_Strix_SemmlA_all.csv};
				\syntunrealplot table [x=time_Strix, y=time_SemmlA] {data/time/SyntcompTF_Strix_SemmlA_real.csv};
				\syntunrealplot table [x=time_Strix, y=time_SemmlA] {data/time/SyntcompTF_Strix_SemmlA_unreal.csv};
			\end{axis}
		\end{tikzpicture}
		
		\begin{tikzpicture}
			\begin{axis}[plot,xmode=log,ymode=log,xlabel=\SemMLa,ylabel=\SemML]
				\axislines
				
				\syntheticplot table [x=time_SemmlA, y=time_Semml] {data/time/Synthetic_SemmlA_Semml_all.csv};
				\syntunrealplot table [x=time_SemmlA, y=time_Semml] {data/time/SyntcompTF_SemmlA_Semml_real.csv};
				\syntunrealplot table [x=time_SemmlA, y=time_Semml] {data/time/SyntcompTF_SemmlA_Semml_unreal.csv};
			\end{axis}
		\end{tikzpicture}
	\end{minipage}
	\caption{
		Scatter plots of the runtimes of \SemML{}, \Strix{}, and \SemMLa{} on \Syntcomp{}.
		A point $(x, y)$ denotes that tool X and tool Y needed $x$ and $y$ seconds, respectively.
		If a point is above/below the diagonal, tool X is faster/slower.
		Plots are on logarithmic scale, dashed diagonals indicate that one tool is twice as fast.
		Timeouts are pushed to the orthogonal dashed line.
		The axes start at 1 second.
	}
	\label{fig:scatterplots}
\end{figure}
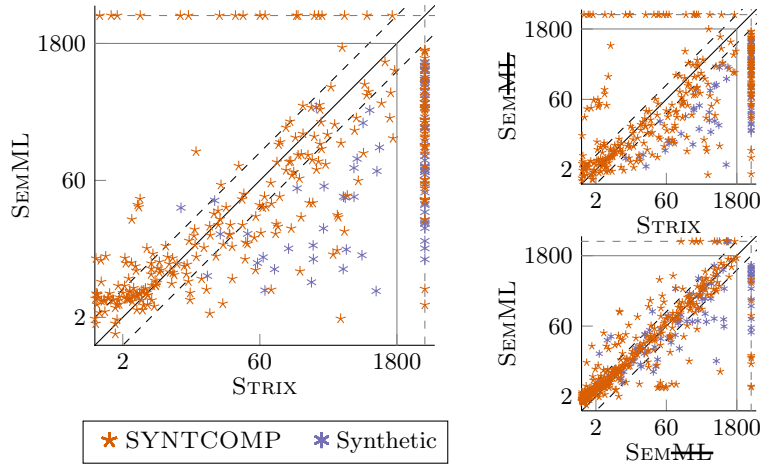

Our main results are summarized in \cref{tbl:results,tbl:speed} and displayed in \cref{fig:scatterplots}.
On \Syntcomp{}, we solve $\StrixSemmlSyntcomptfZeroAllL-\StrixSemmlSyntcomptfZeroAllK=\Diff\StrixSemmlSyntcomptfZeroAllL\StrixSemmlSyntcomptfZeroAllK{}$ instances more than \Strix{} which, considering the difficulty scale of \Syntcomp, marks a major achievement.
Further, we observe significant speed-up compared to \Strix{}, especially as the instances get larger.
For the ratio over entire \Syntcomp{}, we mention that there are about 600 instances that \Strix{} solves (nearly) instantly whereas \SemML{} requires about a second to start the JVM and load ML parameters.

Investigating the unique solves of both tools in more detail, we observe that on many \emph{realizable} families, \SemML{} is able to solve one more instance than \Strix{} within the timeout, sometimes even within a minute.
In particular, there are three families (\texttt{amba\_gr}, \texttt{amba\_decomposed\_lock}, \texttt{collector\_v3}), where two or even more extra instances were solved, marking a major improvement.
The instances only solved by \Strix{} turned out to be mostly from variants of the \texttt{ltl2dba} families.
Here, we conjecture that our guidance takes a \enquote{bad turn}, never reached closure, and thus never reconsidered its steps, while the less guided, broader exploration of \Strix{} has a higher chance of exploring the right parts.

For \emph{unrealizable} formulae, most of our unique solves were of the discussed form of injecting a fault into some arbiter and our targeted exploration was able to localize these faults deep into the state space.
\Strix{}'s unique solves are again mostly unrealizable variants of \texttt{ltl2dba} and \texttt{detector\_unreal} formulae.

On the Synthetic dataset, \SemML{} outperforms \Strix{} by an order of magnitude,
solving all instances \Strix{} is able to solve, and even \StrixSemmlSyntheticZeroAllL{} additional ones.
On the \StrixSemmlSyntheticZeroAllM{} instances that both tools solved, \SemML{} is \StrixSemmlSyntheticZeroAllT{} times faster, and this factor increases to \StrixSemmlSyntheticThreehundredAllT{} when considering the most challenging instances.

\subsection{RQ2: Effects of Machine Learning and Algorithmic Changes}
We proceed to investigate the impact of our algorithmic and engineering changes and ML heuristic individually.
For that matter, we first compare \Strix{} to \SemMLa{} and then proceed with comparing \SemMLa{} to \SemML{}.

\paragraph{\Strix{} vs.\ \SemMLa}
Especially on \Syntcomp{}, a significant part of the improvements is caused by our algorithmic changes.
For example, the game for \texttt{collectorv1\_14} has over 19M states in \Strix{} while \SemML{} keeps it at a manageable 60k states due merging of game states.
This allowed \SemML{} to solve \texttt{collectorv1\_15} as well, whereas \Strix{} could not.
Similarly, we also observed the concrete impact of the BDD improvements as well as the demand-guided exploration.
The former is more prominent for realizable, while the latter shows significant impact mostly (but not exclusively) for unrealizable instances.
Concretely, 24 of \StrixSemmlASyntcomptfZeroUnrealL{} \SemMLa{}'s unique solves of unrealizable instances are fault-injected-arbiters.
Here, our deep, targeted exploration was able to localize the fault, even when only following the exploration score of \Strix{}.
\Strix{}'s broader exploration could not reach these regions despite following the same score, likely because it tried to follow multiple \enquote{leads} at once, due to its global view.
On the Synthetic set, the effects are even more pronounced.
Here, already \SemMLa{} solves all instances that \Strix{} solves and \StrixSemmlASyntheticZeroAllL{} more, with a speed-up of \StrixSemmlASyntheticFiveAllT{}.

\paragraph{\SemMLa{} vs.\ \SemML{}}
Despite our algorithmic changes already yielding significant improvements, ML still adds performance on top.
For \Syntcomp{}, our guidance is able to identify solutions much quicker on several families (e.g.\ \texttt{01-13.tlsf} or \texttt{collector\_v3}), resulting in a speed-up factor of \SemmlASemmlSyntcompThirtyRealT{} (with lower threshold of 30 seconds to focus on complicated instances) and 2 more unique solves on realizable samples.
On unrealizable formulae, \SemML{} solves 7 further instances (mainly from \texttt{full\_arbiter\_unreal}), but fails on 8 instances (mainly from \texttt{round\_robin\_arbiter\_unreal}).
We conjecture that this is due to bad generalization, and due to the fact that exploration towards a deep fault comes with more opportunities for a ML model to \enquote{mess up}, whereas \SemMLa{}'s score is stable.
Overall however, \SemML{} is competitive with \SemMLa{} on \Syntcomp{}, which is positive, considering the discussed structure of \Syntcomp{} and, especially, that our model was not even trained on similar inputs.

Turning our attention to the Synthetic set (in line with the training data), we see that \SemML{} has significant \SemmlASemmlSyntheticZeroAllL{} unique solves compared to \SemmlASemmlSyntheticZeroAllK{} for \SemMLa{} and a speed-up of \SemmlASemmlSyntheticThirtyAllT{} on complicated instances (threshold of 30 seconds).
Based on this, we conjecture that \Syntcomp{} likely is \enquote{out-of-distribution} for our model and, consequently, training with both synthetic as well as \Syntcomp{} samples would add even more improvements for real-world formulae.

\begin{remark}
	To conclude, we stress that our ML models are deliberately kept small, since we need to be able to evaluate them extremely quickly.
	As such, we only considered quite simple features and small models.
	Concretely, the final models for all state classes are comprised of 15 trees of depth 2.
	We also evaluated larger models with more complex and meaningful features.
	These performed significantly better in terms of pure accuracy on the pair classification, but added so much overhead that ultimately fewer instances were solved.
\end{remark}

\section{Conclusion and Future Work}

We presented our tool \SemML, which combines algorithmic and engineering improvements together with tailored machine-learning heuristics to arrive at a highly performant tool for reactive synthesis.
Our experimental evaluation confirms the impact of both our improvements.
In particular, \SemML{} significantly outperforms the state-of-the-art tool \Strix, which dominated the reactive synthesis competition since its first appearance.

For future work, we identify several avenues.
In terms of algorithmic improvements, both the backtracking heuristic and our parity game solver can be significantly improved by a tighter integration with the exploration heuristic.
Further, we intend to implement -- as well as develop new -- state of the art methods for extracting and representing solutions efficiently to practically make use of the solutions we are now able to identify.
For users of the tool, we want to provide a variant of \SemML{} that is also trained on SYNTCOMP formulae, unfair for academic evaluation but useful for actual industrial synthesis. 
We also want to investigate hierarchical oracles, i.e.\ include more accurate (but slower) oracles, which are only consulted when the basic oracle has low confidence.

\clearpage
\paragraph{Data availability statement.}
The models, tools, and scripts to reproduce our experimental evaluation are archived and available at \cite{artifact}.
\bibliographystyle{splncs04}
\bibliography{main}

\begin{thebibliography}{10}
\providecommand{\url}[1]{\texttt{#1}}
\providecommand{\urlprefix}{URL }
\providecommand{\doi}[1]{https://doi.org/#1}

\bibitem{Syntcomp18}
The reactive synthesis competition: {SYNTCOMP} 2018 results.
  http://www.syntcomp.org/syntcomp-2018-results/ (2018),
  \url{http://www.syntcomp.org/syntcomp-2018-results/}

\bibitem{DBLP:journals/tocl/AlurT04}
Alur, R., Torre, S.L.: Deterministic generators and games for ltl fragments.
  {ACM} Trans. Comput. Log.  \textbf{5}(1),  1--25 (2004).
  \doi{10.1145/963927.963928}

\bibitem{DBLP:conf/vstte/BansalGSLVZ22}
Bansal, S., Giacomo, G.D., Stasio, A.D., Li, Y., Vardi, M.Y., Zhu, S.:
  Compositional safety {LTL} synthesis. In: Lal, A., Tonetta, S. (eds.)
  Verified Software. Theories, Tools and Experiments - 14th International
  Conference, {VSTTE} 2022, Trento, Italy, October 17-18, 2022, Revised
  Selected Papers. Lecture Notes in Computer Science, vol. 13800, pp. 1--19.
  Springer (2022). \doi{10.1007/978-3-031-25803-9\_1},
  \url{https://doi.org/10.1007/978-3-031-25803-9\_1}

\bibitem{DBLP:conf/cav/BohyBFJR12}
Bohy, A., Bruy{\`{e}}re, V., Filiot, E., Jin, N., Raskin, J.: Acacia+, a tool
  for {LTL} synthesis. In: Madhusudan, P., Seshia, S.A. (eds.) Computer Aided
  Verification - 24th International Conference, {CAV} 2012, Berkeley, CA, USA,
  July 7-13, 2012 Proceedings. Lecture Notes in Computer Science, vol.~7358,
  pp. 652--657. Springer (2012). \doi{10.1007/978-3-642-31424-7\_45}

\bibitem{Buchi62}
B{\"u}chi, J.: On a decision method in restricted second-order arithmetic. In:
  Nagel, E., Suppes, P., Tarski, A. (eds.) Proceedings of the First
  International Congress on Logic, Methodology, and Philosophy of Science 1960
  (1962)

\bibitem{DBLP:conf/tacas/CadilhacP23}
Cadilhac, M., P{\'{e}}rez, G.A.: Acacia-bonsai: {A} modern implementation of
  downset-based {LTL} realizability. In: Sankaranarayanan, S., Sharygina, N.
  (eds.) Tools and Algorithms for the Construction and Analysis of Systems -
  29th International Conference, {TACAS} 2023, Held as Part of the European
  Joint Conferences on Theory and Practice of Software, {ETAPS} 2022, Paris,
  France, April 22-27, 2023, Proceedings, Part {II}. Lecture Notes in Computer
  Science, vol. 13994, pp. 192--207. Springer (2023).
  \doi{10.1007/978-3-031-30820-8\_14}

\bibitem{Church}
Church, A.: Application of recursive arithmetic to the problem of circuit
  synthesis. Journal of Symbolic Logic  (1963)

\bibitem{DBLP:conf/tacas/CoslerHOS24}
Cosler, M., Hahn, C., Omar, A., Schmitt, F.: Neurosynt: {A} neuro-symbolic
  portfolio solver for reactive synthesis. In: Finkbeiner, B., Kov{\'{a}}cs, L.
  (eds.) Tools and Algorithms for the Construction and Analysis of Systems -
  30th International Conference, {TACAS} 2024, Held as Part of the European
  Joint Conferences on Theory and Practice of Software, {ETAPS} 2024,
  Luxembourg City, Luxembourg, April 6-11, 2024, Proceedings, Part {III}.
  Lecture Notes in Computer Science, vol. 14572, pp. 45--67. Springer (2024).
  \doi{10.1007/978-3-031-57256-2\_3}

\bibitem{DBLP:conf/iclr/CoslerSHF23}
Cosler, M., Schmitt, F., Hahn, C., Finkbeiner, B.: Iterative circuit repair
  against formal specifications. In: The Eleventh International Conference on
  Learning Representations, {ICLR} 2023, Kigali, Rwanda, May 1-5, 2023.
  OpenReview.net (2023), \url{https://openreview.net/forum?id=SEcSahl0Ql}

\bibitem{specificationPatterns}
Dwyer, M.B., Avrunin, G.S., Corbett, J.C.: Property specification patterns for
  finite-state verification. In: Proceedings of the Second Workshop on Formal
  Methods in Software Practice. p. 7–15. FMSP '98, Association for Computing
  Machinery, New York, NY, USA (1998). \doi{10.1145/298595.298598}

\bibitem{DBLP:conf/tacas/Ehlers11}
Ehlers, R.: Unbeast: Symbolic bounded synthesis. In: Abdulla, P.A., Leino,
  K.R.M. (eds.) Tools and Algorithms for the Construction and Analysis of
  Systems - 17th International Conference, {TACAS} 2011, Held as Part of the
  Joint European Conferences on Theory and Practice of Software, {ETAPS} 2011,
  Saarbr{\"{u}}cken, Germany, March 26-April 3, 2011. Proceedings. Lecture
  Notes in Computer Science, vol.~6605, pp. 272--275. Springer (2011).
  \doi{10.1007/978-3-642-19835-9\_25}

\bibitem{DBLP:journals/sttt/EsparzaKRS22}
Esparza, J., Kret{\'{\i}}nsk{\'{y}}, J., Raskin, J., Sickert, S.: From linear
  temporal logic and limit-deterministic b{\"{u}}chi automata to deterministic
  parity automata. Int. J. Softw. Tools Technol. Transf.  \textbf{24}(4),
  635--659 (2022). \doi{10.1007/s10009-022-00663-1}

\bibitem{DBLP:journals/jacm/EsparzaKS20}
Esparza, J., Kret{\'{\i}}nsk{\'{y}}, J., Sickert, S.: A unified translation of
  linear temporal logic to {\(\omega\)}-automata. J. {ACM}  \textbf{67}(6),
  33:1--33:61 (2020). \doi{10.1145/3417995}

\bibitem{DBLP:conf/cav/FaymonvilleFT17}
Faymonville, P., Finkbeiner, B., Tentrup, L.: Bosy: An experimentation
  framework for bounded synthesis. In: Majumdar, R., Kuncak, V. (eds.) Computer
  Aided Verification - 29th International Conference, {CAV} 2017, Heidelberg,
  Germany, July 24-28, 2017, Proceedings, Part {II}. Lecture Notes in Computer
  Science, vol. 10427, pp. 325--332. Springer (2017).
  \doi{10.1007/978-3-319-63390-9\_17}

\bibitem{DBLP:journals/corr/Jacobs016}
Jacobs, S., Klein, F., Schirmer, S.: A high-level {LTL} synthesis format:
  {TLSF} v1.1. In: Piskac, R., Dimitrova, R. (eds.) Proceedings Fifth Workshop
  on Synthesis, SYNT@CAV 2016, Toronto, Canada, July 17-18, 2016. {EPTCS},
  vol.~229, pp. 112--132 (2016). \doi{10.4204/EPTCS.229.10}

\bibitem{DBLP:journals/corr/abs-2206-00251}
Jacobs, S., P{\'{e}}rez, G.A., Abraham, R., Bruy{\`{e}}re, V., Cadilhac, M.,
  Colange, M., Delfosse, C., van Dijk, T., Duret{-}Lutz, A., Faymonville, P.,
  Finkbeiner, B., Khalimov, A., Klein, F., Luttenberger, M., Meyer, K.J.,
  Michaud, T., Pommellet, A., Renkin, F., Schlehuber{-}Caissier, P., Sakr, M.,
  Sickert, S., Staquet, G., Tamines, C., Tentrup, L., Walker, A.: The reactive
  synthesis competition {(SYNTCOMP):} 2018-2021. CoRR  \textbf{abs/2206.00251}
  (2022). \doi{10.48550/arXiv.2206.00251}

\bibitem{DBLP:conf/fmcad/JobstmannB06}
Jobstmann, B., Bloem, R.: Optimizations for {LTL} synthesis. In: Formal Methods
  in Computer-Aided Design, 6th International Conference, {FMCAD} 2006, San
  Jose, California, USA, November 12-16, 2006, Proceedings. pp. 117--124.
  {IEEE} Computer Society (2006). \doi{10.1109/FMCAD.2006.22}

\bibitem{DBLP:conf/cav/KretinskyE12}
Kret{\'{\i}}nsk{\'{y}}, J., Esparza, J.: Deterministic automata for the (f,
  g)-fragment of {LTL}. In: Madhusudan, P., Seshia, S.A. (eds.) Computer Aided
  Verification - 24th International Conference, {CAV} 2012, Berkeley, CA, USA,
  July 7-13, 2012 Proceedings. Lecture Notes in Computer Science, vol.~7358,
  pp. 7--22. Springer (2012). \doi{10.1007/978-3-642-31424-7\_7}

\bibitem{DBLP:conf/atva/KretinskyMM19}
Kret{\'{\i}}nsk{\'{y}}, J., Manta, A., Meggendorfer, T.: Semantic labelling and
  learning for parity game solving in {LTL} synthesis. In: Chen, Y., Cheng, C.,
  Esparza, J. (eds.) Automated Technology for Verification and Analysis - 17th
  International Symposium, {ATVA} 2019, Taipei, Taiwan, October 28-31, 2019,
  Proceedings. Lecture Notes in Computer Science, vol. 11781, pp. 404--422.
  Springer (2019). \doi{10.1007/978-3-030-31784-3\_24}

\bibitem{CAV23}
Kret{\'{\i}}nsk{\'{y}}, J., Meggendorfer, T., Prokop, M., Rieder, S.: Guessing
  winning policies in {LTL} synthesis by semantic learning. In: Enea, C., Lal,
  A. (eds.) Computer Aided Verification - 35th International Conference, {CAV}
  2023, Paris, France, July 17-22, 2023, Proceedings, Part {I}. Lecture Notes
  in Computer Science, vol. 13964, pp. 390--414. Springer (2023).
  \doi{10.1007/978-3-031-37706-8\_20}

\bibitem{DBLP:conf/atva/KretinskyMS18}
Kret{\'{\i}}nsk{\'{y}}, J., Meggendorfer, T., Sickert, S.: Owl: {A} library for
  {\(\omega\)}-words, automata, and {LTL}. In: Lahiri, S.K., Wang, C. (eds.)
  Automated Technology for Verification and Analysis - 16th International
  Symposium, {ATVA} 2018, Los Angeles, CA, USA, October 7-10, 2018,
  Proceedings. Lecture Notes in Computer Science, vol. 11138, pp. 543--550.
  Springer (2018). \doi{10.1007/978-3-030-01090-4\_34}

\bibitem{DBLP:conf/cav/KretinskyMSZ18}
Kret{\'{\i}}nsk{\'{y}}, J., Meggendorfer, T., Sickert, S., Ziegler, C.:
  Rabinizer 4: From {LTL} to your favourite deterministic automaton. In:
  Chockler, H., Weissenbacher, G. (eds.) Computer Aided Verification - 30th
  International Conference, {CAV} 2018, Held as Part of the Federated Logic
  Conference, FloC 2018, Oxford, UK, July 14-17, 2018, Proceedings, Part {I}.
  Lecture Notes in Computer Science, vol. 10981, pp. 567--577. Springer (2018).
  \doi{10.1007/978-3-319-96145-3\_30}

\bibitem{DBLP:conf/cav/KupfermanPV06}
Kupferman, O., Piterman, N., Vardi, M.Y.: Safraless compositional synthesis.
  In: Ball, T., Jones, R.B. (eds.) Computer Aided Verification, 18th
  International Conference, {CAV} 2006, Seattle, WA, USA, August 17-20, 2006,
  Proceedings. Lecture Notes in Computer Science, vol.~4144, pp. 31--44.
  Springer (2006). \doi{10.1007/11817963\_6}

\bibitem{jbdd}
Meggendorfer, T.: {JBDD}: A java {BDD} library.
  \url{https://github.com/incaseoftrouble/jbdd} (2017)

\bibitem{PET}
Meggendorfer, T., Weininger, M.: Playing games with your pet: Extending the
  partial exploration tool to stochastic games (2024),
  \url{https://arxiv.org/abs/2405.03885}

\bibitem{DBLP:conf/cav/MeyerSL18}
Meyer, P.J., Sickert, S., Luttenberger, M.: Strix: Explicit reactive synthesis
  strikes back! In: Chockler, H., Weissenbacher, G. (eds.) Computer Aided
  Verification - 30th International Conference, {CAV} 2018, Held as Part of the
  Federated Logic Conference, FloC 2018, Oxford, UK, July 14-17, 2018,
  Proceedings, Part {I}. Lecture Notes in Computer Science, vol. 10981, pp.
  578--586. Springer (2018). \doi{10.1007/978-3-319-96145-3\_31}

\bibitem{scikit-learn}
Pedregosa, F., Varoquaux, G., Gramfort, A., Michel, V., Thirion, B., Grisel,
  O., Blondel, M., Prettenhofer, P., Weiss, R., Dubourg, V., Vanderplas, J.,
  Passos, A., Cournapeau, D., Brucher, M., Perrot, M., Duchesnay, E.:
  Scikit-learn: Machine learning in {P}ython. Journal of Machine Learning
  Research  \textbf{12},  2825--2830 (2011)

\bibitem{DBLP:conf/lics/Piterman06}
Piterman, N.: From nondeterministic buchi and streett automata to deterministic
  parity automata. In: 21th {IEEE} Symposium on Logic in Computer Science
  {(LICS} 2006), 12-15 August 2006, Seattle, WA, USA, Proceedings. pp.
  255--264. {IEEE} Computer Society (2006). \doi{10.1109/LICS.2006.28}

\bibitem{DBLP:conf/vmcai/PitermanPS06}
Piterman, N., Pnueli, A., Sa'ar, Y.: Synthesis of reactive(1) designs. In:
  Emerson, E.A., Namjoshi, K.S. (eds.) Verification, Model Checking, and
  Abstract Interpretation, 7th International Conference, {VMCAI} 2006,
  Charleston, SC, USA, January 8-10, 2006, Proceedings. Lecture Notes in
  Computer Science, vol.~3855, pp. 364--380. Springer (2006).
  \doi{10.1007/11609773\_24}

\bibitem{Pnueli77}
Pnueli, A.: The temporal logic of programs. In: 18th Annual Symposium on
  Foundations of Computer Science, Providence, Rhode Island, USA, 31 October -
  1 November 1977. pp. 46--57. {IEEE} Computer Society (1977).
  \doi{10.1109/SFCS.1977.32}

\bibitem{DBLP:conf/icalp/PnueliR89}
Pnueli, A., Rosner, R.: On the synthesis of an asynchronous reactive module.
  In: Ausiello, G., Dezani{-}Ciancaglini, M., Rocca, S.R.D. (eds.) Automata,
  Languages and Programming, 16th International Colloquium, ICALP89, Stresa,
  Italy, July 11-15, 1989, Proceedings. Lecture Notes in Computer Science,
  vol.~372, pp. 652--671. Springer (1989). \doi{10.1007/BFb0035790}

\bibitem{artifact}
Prokop, M.: Artifact for "semml: Enhancing automata-theoretic ltl synthesis
  with machine learning" (Jan 2025). \doi{10.5281/zenodo.14587814},
  \url{https://doi.org/10.5281/zenodo.14587814}

\bibitem{DBLP:journals/fmsd/RenkinSDP22}
Renkin, F., Schlehuber{-}Caissier, P., Duret{-}Lutz, A., Pommellet, A.:
  Dissecting ltlsynt. Formal Methods Syst. Des.  \textbf{61}(2),  248--289
  (2022). \doi{10.1007/S10703-022-00407-6}

\bibitem{RealizabilityModuloTheories}
Rodr{\'i}guez, A., S{\'a}nchez, C.: Boolean abstractions for realizability
  modulo theories. In: Enea, C., Lal, A. (eds.) Computer Aided Verification.
  pp. 305--328. Springer Nature Switzerland, Cham (2023)

\bibitem{DBLP:conf/focs/Safra88}
Safra, S.: On the complexity of omega-automata. In: 29th Annual Symposium on
  Foundations of Computer Science, White Plains, New York, USA, 24-26 October
  1988. pp. 319--327. {IEEE} Computer Society (1988).
  \doi{10.1109/SFCS.1988.21948}

\bibitem{DBLP:conf/fossacs/Schewe09}
Schewe, S.: Tighter bounds for the determinisation of b{\"{u}}chi automata. In:
  de~Alfaro, L. (ed.) Foundations of Software Science and Computational
  Structures, 12th International Conference, {FOSSACS} 2009, Held as Part of
  the Joint European Conferences on Theory and Practice of Software, {ETAPS}
  2009, York, UK, March 22-29, 2009. Proceedings. Lecture Notes in Computer
  Science, vol.~5504, pp. 167--181. Springer (2009).
  \doi{10.1007/978-3-642-00596-1\_13}

\bibitem{DBLP:conf/nips/SchmittHRF21}
Schmitt, F., Hahn, C., Rabe, M.N., Finkbeiner, B.: Neural circuit synthesis
  from specification patterns. In: Ranzato, M., Beygelzimer, A., Dauphin, Y.N.,
  Liang, P., Vaughan, J.W. (eds.) Advances in Neural Information Processing
  Systems 34: Annual Conference on Neural Information Processing Systems 2021,
  NeurIPS 2021, December 6-14, 2021, virtual. pp. 15408--15420 (2021),
  \url{https://proceedings.neurips.cc/paper/2021/hash/8230bea7d54bcdf99cdfe85cb07313d5-Abstract.html}

\bibitem{Sickert_16_LDBA}
Sickert, S., Esparza, J., Jaax, S., K{\v{r}}et{\'{i}}nsk{\'{y}}, J.:
  Limit-deterministic b{\"{u}}chi automata for linear temporal logic. In:
  Chaudhuri, S., Farzan, A. (eds.) Computer Aided Verification - 28th
  International Conference, {CAV} 2016, Toronto, ON, Canada, July 17-23, 2016,
  Proceedings, Part {II}. Lecture Notes in Computer Science, vol.~9780, pp.
  312--332. Springer (2016). \doi{10.1007/978-3-319-41540-6\_17}

\bibitem{sickert_meyer_philipp_modernizing_2021}
Sickert, Meyer~Philipp, S.: Modernizing {Strix} (2021),
  \url{https://www7.in.tum.de/~sickert/publications/MeyerS21.pdf}

\bibitem{DBLP:conf/stacs/Thomas95}
Thomas, W.: On the synthesis of strategies in infinite games. In: Mayr, E.W.,
  Puech, C. (eds.) {STACS} 95, 12th Annual Symposium on Theoretical Aspects of
  Computer Science, Munich, Germany, March 2-4, 1995, Proceedings. Lecture
  Notes in Computer Science, vol.~900, pp. 1--13. Springer (1995).
  \doi{10.1007/3-540-59042-0\_57}

\bibitem{DBLP:journals/acta/TomitaUSHY17}
Tomita, T., Ueno, A., Shimakawa, M., Hagihara, S., Yonezaki, N.: Safraless
  {LTL} synthesis considering maximal realizability. Acta Informatica
  \textbf{54}(7),  655--692 (2017). \doi{10.1007/S00236-016-0280-3}

\end{thebibliography}

\iftoggle{arxiv}{
\newpage
\appendix
\section{Appendix}

\subsection{Details on Feature Elimination and Comparing Different Models} \label{app:model_experiments}

We provide further details on the experiments from \cref{sec:Training}, Training, where we perform recursive feature elimination on five different model types.
Recall that we are looking for a good implementation for the pair classifier of our exploration heuristic.
Thus, we compare various binary classification models such as Neural networks, linear and kernel SVMs, random forests, and gradient boosted trees.
As each model type possibly requires their own set of features, we perform an individual RFE for each of them.
Additionally, each state class might require different features, thus we perform individual RFEs for them as well.
We call these state classes Env0, Env1, Sys0, Sys1, where the prefix denotes the player, and the number denotes whether acceptance information is ignored ($0$ meaning no, $1$ being yes).

Our RFE works slightly different, as we operate on pairs of edges and thus every feature appears twice.
For the tree models, each feature even appears three times, as we include the pointwise difference as well.
Thus, when we obtain the feature importances, we have to take the maximum for each feature over all its appearances.
While feature importances are directly provided by linear SVMs and tree models, for kernel SVMs and Neural networks, we resort to \textit{permutation importance}, which is also provided by \texttt{sklearn}.

The results can be observed in \cref{fig:rfe}.
For each model type, the model we considered most suitable is highlighted.
Note that this isn't necessarily the model with the best accuracy, due to the advice being a very performance critical component of the overall procedure.
Thus, we might prefer a model with a little less accuracy if way fewer features are required to achieve it.

\begin{figure}[tp]
	\label{fig:rfe}
	\begin{minipage}{0.49\textwidth}
		\includegraphics[width=\textwidth]{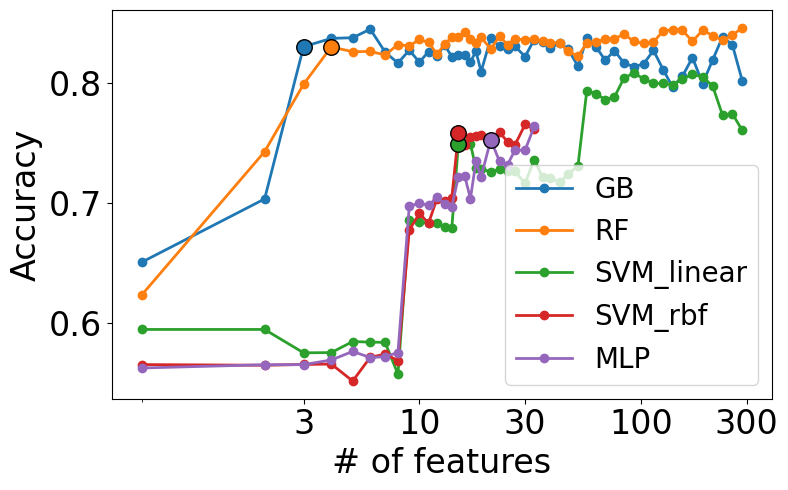}
		
		\includegraphics[width=\textwidth]{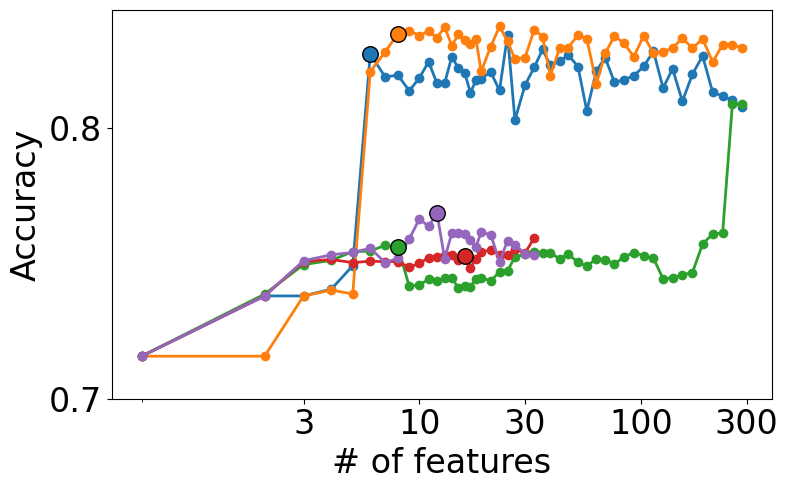}
	\end{minipage}
	\begin{minipage}{0.49\textwidth}
		\includegraphics[width=\textwidth]{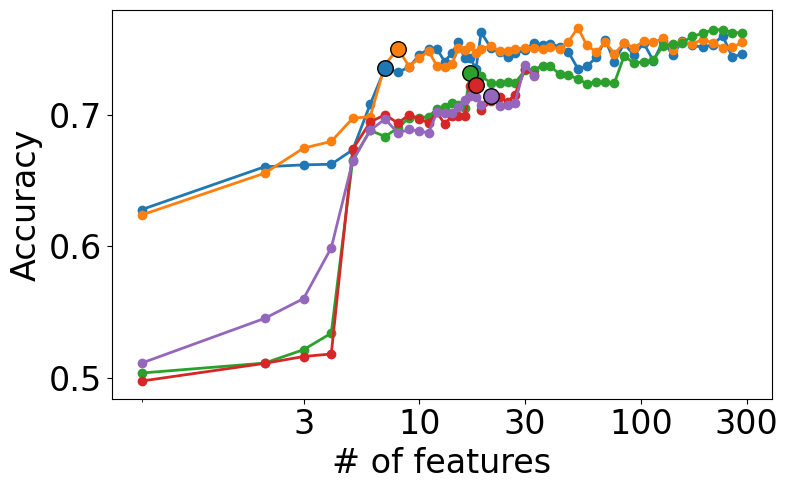}
		
		\includegraphics[width=\textwidth]{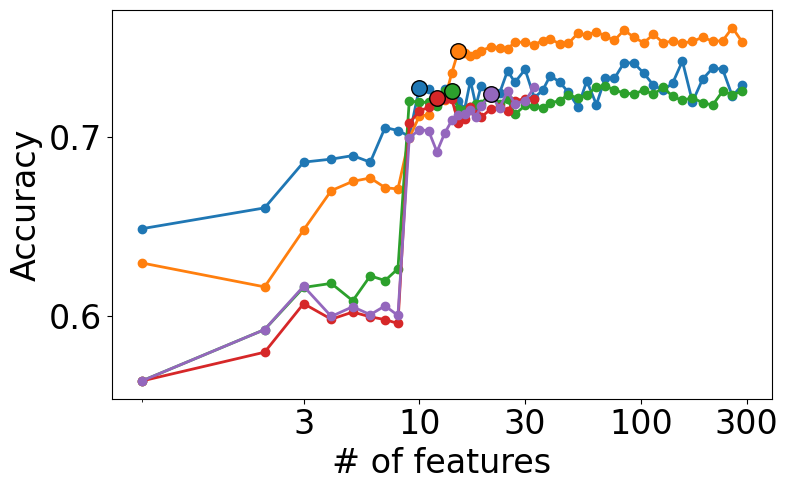}
	\end{minipage}
	
	\caption{
		Details on the feature elimination experiments for model type and state class individually.
		Specifically, we have the plots for Env0 (top left), Env1 (bottom left), Sys0 (top right), and Sys1 (bottom right).
		For every state class and model type, we highlight the \enquote{best} model, that moves on to the second evaluation round.
	}
	
\end{figure}

\subsection{State wise scoring}

Pure accuracy on pair classification does not necessarily translate to good performance as exploration heuristic (although it certainly helps)
Consider a state with 4 edges $e_1,\ldots,e_4$ where $e_1$ is by far the best option, and $e_2,e_3,e_4$ are roughly equal but clearly worse than $e_1$.
Then a model which only classifies the pairs containing $e_1$ correctly is much more useful than a model that only classifies all other pairs correctly, despite both having an accuracy of $0.5$.
In other words, it is much more important to correctly classify pairs containing the best edges than correctly distinguishing between the 5th and 6th best edge.

For that reason, we propose a state wise score which evaluates the models w.r.t.\ the quality of the advice they would give for entire states.
Let $s$ be some system state, $E(s)$ its outgoing edges, and $\textit{gt}(e)$ the ground truth value of edge $e$.
The score for predicting edge $e$ for state $s$ is given by:
\begin{equation*}
	\textit{score}(e) = \frac{\textit{gt}(e)}{\max\limits_{e'\in E(s)} \textit{gt}(e')}
\end{equation*} 
Intuitively, if the best edge (the one with the highest ground truth value) is predicted, the model obtains a score of 1 and for every worse edge a respectively scaled number between $0$ and $1$.
Note that for environment states, where the ground truth values are between $0$ and $-1$, the maximum in the denominator becomes a minimum.
The interpretation stays the same, however.
Also note that the score only exists if a player wins their respective state (otherwise we obtain $0/0$).
Therefore, states of one player that are won by the other player are excluded.
Further, states where every edge has the same ground truth value (in particular, states with only one edge), are excluded as well, as these would always yield a score of $1$ and thus inflate the overall score.

In \cref{tbl:stateScores} we depict the average state scores of all models on a test set of synthetic instances (which is neither the train, nor the validation, nor the large evaluation set from the main body).
We consider each state class separately.
The specific model instance used for every model type is marked in \cref{fig:rfe}.
As baselines, we include the scores of a random advice, as well as advice only based on the exploration score of \Strix{}, i.e.\ the advice employed in \SemMLa{}.

\begin{table}[t]
	\centering
	\caption{State scores of every model type split by state class.}
	\setlength{\tabcolsep}{5pt}
	\begin{tabular}{lccccccc}
		\label{tbl:stateScores} & rand  & \Strix{} & lin-SVM & rbf-SVM &  MLP  &  RF   &  GBT  \\
		\midrule
		Env. Class 0            & 0.503 &  0.614   &  0.777  &  0.769  & 0.753 & 0.831 & 0.807 \\
		Env. Class 1            & 0.517 &  0.745   &  0.882  &  0.885  & 0.895 & 0.936 & 0.927 \\
		Sys. Class 0            & 0.496 &  0.628   &  0.833  &  0.805  & 0.856 & 0.817 & 0.836 \\
		Sys. Class 1            & 0.498 &  0.590   &  0.768  &  0.750  & 0.746 & 0.787 & 0.748
	\end{tabular}
\end{table}

There are a few things to note here:
Firstly, as expected, every ML model heavily outscores the baseline of \Strix{}.
Secondly, GBT gets outscored by a few points on 3/4 classes, especially by RF.
However, GBT uses consistently less features to achieve its scores which is why we deemed it more suitable for practical purposes.
Finally, random choice yields roughly 0.5, indicating the score mimics accuracy

\subsection{Full Detail Comparison of \Strix{}, \SemMLa{}, and \SemML{}}

For a full detain comparison, we employ the four-figure score suggested by \cite{PET}, which allows for a concise pairwise comparison of two tools and is written as $t[m{+}k/l]$.
For comparing two tools $A$ and $B$, we first identify the set of benchmarks $M$ that both tools were able to solve within their constraints.
Then, $t$ is the geometric mean of the ratios $\textit{time}_A(I)/\textit{time}_B(I)$ for $I \in M$, where $\textit{time}_T(I)$ is the time required for tool $T$ to solve instance $I$.
Further, $m =|M|$, $k$ is the number of instances that only $A$ was able to solve and $l$ vice versa.
Intuitively, $t<1$ means tool $A$ solves the instances of $M$ faster.
This usually coincides with cases where $k>l$, i.e.\ tool $A$ solves more instances that tool $B$ cannot solve.
However, in cases where $t>1$ but $k \gg l$ one may still prefer tool $A$ as it solves more instances overall.

\Cref{tbl:syntcompScores,tbl:syntheticScores} show all scores we computed for our evaluation on \Syntcomp{} data and Synthetic data, respectively.

\begin{table}[tp]
\centering
\caption{Scores of all pairings on the \Syntcomp{} dataset separated by realizability, and lower cutoff.
The scores of a column labelled x, only contain samples where at least one tool required more than x seconds.
Intuitively, the higher this lower cutoff is, the more a score focusses on complex samples.
Note that the number of unique solves, is independent of the lower cutoff.
}
\scriptsize
\begin{tabular}{lccccc}
	\label{tbl:syntcompScores}
	\Strix{}/\SemML &                   0                   &                  2                   &                   5                   &                   30                    &                      300                      \\
	\midrule
	Real                                        &  \StrixSemmlSyntcomptfZeroRealScore   &  \StrixSemmlSyntcomptfTwoRealScore   &  \StrixSemmlSyntcomptfFiveRealScore   &  \StrixSemmlSyntcomptfThirtyRealScore   &  \StrixSemmlSyntcomptfThreehundredRealScore   \\
	Uneal                                       & \StrixSemmlSyntcomptfZeroUnrealScore  & \StrixSemmlSyntcomptfTwoUnrealScore  & \StrixSemmlSyntcomptfFiveUnrealScore  & \StrixSemmlSyntcomptfThirtyUnrealScore  & \StrixSemmlSyntcomptfThreehundredUnrealScore  \\
	All                                         &   \StrixSemmlSyntcomptfZeroAllScore   &   \StrixSemmlSyntcomptfTwoAllScore   &   \StrixSemmlSyntcomptfFiveAllScore   &   \StrixSemmlSyntcomptfThirtyAllScore   &   \StrixSemmlSyntcomptfThreehundredAllScore   \\
	                                            &                                       &                                      &                                       &                                         &                                               \\
	                                            &                                       &                                      &                                       &                                         &                                               \\
	\Strix{}/\SemMLa                            &                   0                   &                  2                   &                   5                   &                   30                    &                      300                      \\
	\midrule
	Real                                        &  \StrixSemmlASyntcomptfZeroRealScore  &  \StrixSemmlASyntcomptfTwoRealScore  &  \StrixSemmlASyntcomptfFiveRealScore  &  \StrixSemmlASyntcomptfThirtyRealScore  &  \StrixSemmlASyntcomptfThreehundredRealScore  \\
	Uneal                                       & \StrixSemmlASyntcomptfZeroUnrealScore & \StrixSemmlASyntcomptfTwoUnrealScore & \StrixSemmlASyntcomptfFiveUnrealScore & \StrixSemmlASyntcomptfThirtyUnrealScore & \StrixSemmlASyntcomptfThreehundredUnrealScore \\
	All                                         &  \StrixSemmlASyntcomptfZeroAllScore   &  \StrixSemmlASyntcomptfTwoAllScore   &  \StrixSemmlASyntcomptfFiveAllScore   &  \StrixSemmlASyntcomptfThirtyAllScore   &  \StrixSemmlASyntcomptfThreehundredAllScore   \\
	                                            &                                       &                                      &                                       &                                         &                                               \\
	                                            &                                       &                                      &                                       &                                         &                                               \\
	\SemMLa{}/\SemML                            &                   0                   &                  2                   &                   5                   &                   30                    &                      300                      \\
	\midrule
	Real                                        &  \SemmlASemmlSyntcomptfZeroRealScore  &  \SemmlASemmlSyntcomptfTwoRealScore  &  \SemmlASemmlSyntcomptfFiveRealScore  &  \SemmlASemmlSyntcomptfThirtyRealScore  &  \SemmlASemmlSyntcomptfThreehundredRealScore  \\
	Uneal                                       & \SemmlASemmlSyntcomptfZeroUnrealScore & \SemmlASemmlSyntcomptfTwoUnrealScore & \SemmlASemmlSyntcomptfFiveUnrealScore & \SemmlASemmlSyntcomptfThirtyUnrealScore & \SemmlASemmlSyntcomptfThreehundredUnrealScore \\
	All                                         &  \SemmlASemmlSyntcomptfZeroAllScore   &  \SemmlASemmlSyntcomptfTwoAllScore   &  \SemmlASemmlSyntcomptfFiveAllScore   &  \SemmlASemmlSyntcomptfThirtyAllScore   &  \SemmlASemmlSyntcomptfThreehundredAllScore
\end{tabular}
\end{table}

\begin{table}[tp]
\caption{Scores of all pairings on the Synthetic dataset separated by realizability}
\centering
\setlength{\tabcolsep}{10pt}
\begin{tabular}{lccc}
	\label{tbl:syntheticScores}
	Synthetic &           \Strix{}/\SemML           &           \Strix{}/\SemMLa           &           \SemMLa{}/\SemML           \\
	\midrule
	Real                                   &  \StrixSemmlSyntheticZeroRealScore   &  \StrixSemmlASyntheticZeroRealScore  &  \SemmlASemmlSyntheticZeroRealScore  \\
	Unreal                                 &  \StrixSemmlSyntheticZeroUnrealScore & \StrixSemmlASyntheticZeroUnrealScore & \SemmlASemmlSyntheticZeroUnrealScore \\
	All                                    &  \StrixSemmlSyntheticZeroAllScore    &  \StrixSemmlASyntheticZeroAllScore   &  \SemmlASemmlSyntheticZeroAllScore
\end{tabular}
\end{table}

\subsection{Comparison with \Neurosynt}\label{app:neurosynt}
In this section we provide a comparison of \SemML{} and \Neurosynt{} \cite{DBLP:conf/tacas/CoslerHOS24}, which in principle is an even stronger tool than \Strix{} (just in terms of samples solved).
This, however, is not too surprising, given that \Neurosynt{} is a portfolio solver that runs several tools (e.g.\ a neural solver based on \cite{DBLP:conf/nips/SchmittHRF21,DBLP:conf/iclr/CoslerSHF23} and in particular: \Strix{} itself) in parallel and reports the first solution found by any tool.
However, due to the portfolio nature, a competitive comparison of \SemML{} and \Neurosynt{} is neither exactly fair nor meaningful in evaluating the approach of \SemML{}.
It makes much more sense to compare \SemML{} to \enquote{the symbolic component} of \Neurosynt{} (which is essentially covered by \Strix{}) which is why we opted for that comparison being the main experiment of this paper.
Nevertheless, comparing \SemML{} to \Neurosynt{} is interesting on its own which is why we provide this supplementary experiment.

\paragraph{Experimental Setup}
When setting up \Neurosynt{} on our machine, we unfortunately faced technical barriers that we were unable to resolve, even in correspondence with the authors.
Thus, together with the authors of \Neurosynt{}, we concluded on the following setup that is feasibly while also providing meaningful insights:
Since the main difference between \Neurosynt{} and \Strix{} is the neural solver from \cite{DBLP:conf/nips/SchmittHRF21,DBLP:conf/iclr/CoslerSHF23}, we mainly need results for said neural solver on our benchmarks.
These were kindly provided by the authors of \cite{DBLP:conf/tacas/CoslerHOS24} (although only for \Syntcomp{}, as the synthetic data is not in the assumption-guarantee format that the neural solver requires).
We then combine their results with our results for \Strix{} to obtain a good proxy for \Neurosynt{}.
Since the runtime does not mean much (due to different machines, the parallelization of \Neurosynt{} and the \enquote{guess and check} nature of the neural solver), we focus solely on instances solved.
The results can be observed in \Cref{tbl:resultsNeuroSynt}.

\begin{table}[t]
	\caption{
		Comparison of \SemML{} to \Neurosynt{} (with \Strix{} as only symbolic solver) as well as \SemML{} and only the neural solver on \Syntcomp{} data.
	} \label{tbl:resultsNeuroSynt}
	\centering
	\setlength{\tabcolsep}{4pt}
	\setlength{\extrarowheight}{3pt}
	\renewcommand{\arraystretch}{1.1}
	\begin{minipage}{0.49\textwidth}
		\centering
		\begin{tabular}{lccc}
			\multicolumn{2}{c}{\multirow{2}{*}{\Syntcomp}}          &                    \multicolumn{2}{c}{\Neurosynt}                    \\
			&          &            solved             &             unsolved             \\
			\midrule
			\multirow{2}{*}{\rotatebox[origin=c]{90}{\SemML}} &  solved  & 966 &  34   \\
			& unsolved & 35 & 76
		\end{tabular}
	\end{minipage}
	\begin{minipage}{0.49\textwidth}
		\centering
		\begin{tabular}{lccc}
			\multicolumn{2}{c}{\multirow{2}{*}{\Syntcomp}}        &                   \multicolumn{2}{c}{Neural solver}                   \\
			&          &            solved            &            unsolved             \\
			\midrule
			\multirow{2}{*}{\rotatebox[origin=c]{90}{\SemML}} &  solved  & 341 &  659   \\
			& unsolved & 8 & 103
		\end{tabular}
	\end{minipage}
\end{table}

\paragraph{Discussion}
First of all, it is worth noticing that a big part of the performance of \Neurosynt{} comes from \Strix{} (note the 659 samples that are not solved by the neural solver but solved by \SemML), which is why a large part of the analysis overlaps with \Cref{sec:experiments}.
Nevertheless, there are some interesting aspects to discuss. 
First of all, there are 8 samples that are solved by the neural solver that \SemML{} could not solve.
All but one of these are from the family \texttt{full\_arbiter\_unreal} with large parameters.
We conjecture that the neural solver has learned the \enquote{structure} of this family and thus is able to guess correct solutions \enquote{independent} of the parameters.
Since this is a family where \SemML{} performed quite well compared to \Strix{} (due to our targeted exploration that hones in on the one fault injected in the specification), the neural solver not only creates new unique solves but also solves the instances that \Strix{} could not solve.
In fact, this family is where most of the difference between \Neurosynt{} and \Strix{} stem from.
Specifically, 19 out of the 23 unique solves of \Neurosynt{} over \Strix{} are from this family.

Second, while \Neurosynt{} does end up solving one more sample, it is worth noting that \SemML{} still has 34 unique solves over the portfolio of \Strix{} and the neural solver.
This again mainly stems from \SemML{}'s advantage over \Strix{} and shows that \Neurosynt{} certainly would profit from incorporating \SemML{} in their portfolio.

\subsection{Discussion of Features} \label{app:features}
In this section, we want to provide some more insight on the features and specifically which ones made the final cut.
As already hinted in the main body, our features come with several switches and subtly different variants.
For a concrete example, one of our features is named
\begin{center}
	\texttt{E\_R\_U\_S\_flat\_S\_R\_F\_A\_id\_trueness\_all\_prop\_state\_bool-d}.
\end{center}
Thus, we believe that providing full detail on every switch is not particularly helpful.
Instead we report on the most frequent values of the most important switches and explain their meaning.
This requires some understanding of the semantic labelling and the underlying automaton construction.
For that we refer to \cite{sickert_meyer_philipp_modernizing_2021} for a rough overview.

\paragraph{Formula Features}
First, the switch deciding the formula feature (e.g.\ \enquote{trueness} in the above) is equally distributed over features representing \enquote{formula-sat-difficulty},  \enquote{formula-controllability}, or \enquote{formula-complexity}.
For \enquote{formula-sat-difficulty} the most prominent instance is a variant of trueness, where the opponents variables are quantified.
Recall that the trueness of a formula is the number of satisfying assignments divided by the number of total assignments (swapping out temporal operators for new propositional variables).
In the quantified variant, an assignment is only counted towards the numerator if it satisfies the formula under all (resp.\ one) assignments for the opponents variables.
This introduces an aspect of formula controllability as well.

For \enquote{formula-controllability} the most frequent instances were either the above quantified trueness, or a quantitative measure focused only on controllability inspired by \cite{CAV23}.
In essence, we define an inductive function that for every operator yields the controllability based on the controllability of its operands.
E.g.\ the formula $\ltlGlobally \phi$ is as controllable as the operand $\phi$ is controllable.
Note that this does not detect contradictions.
In particular, the formula $a \land \neg a$, where $a$ is owned by the system, would be considered very controllable, as the system has full control over both operands of the and operator. 
However, since many of these (simple) contradictions are removed by preprocessing and the normalization procedure of the automaton construction, the feature (evidently) performs reasonably well.

Finally, for \enquote{formula-complexity} the most common measure is the number of temporal operators in the formula followed by the number of top level disjuncts as a close second.
While it is interesting that the number of temporal operators reliably beat e.g.\ the height of the syntax tree, we would not interpret too much into this particular result. 

\paragraph{Other switches}
First, there is a switch that denotes whether the feature represents a value for the first or second edge of the pair, or the difference between both edges (Recall that we included this specifically for tree models as they cannot compute differences by themselves).
This worked as intended, as over 60\% of the features requested by our final models (which are gradient boosted trees) are difference features.
The remaining 40\% are features specific to the first or second edge.

Further, around 33\% of features use the statewide normalization whereas the remaining 66\% do not.
While we would have expected it to trend the other way, it shows that a mixture of both normalized (representing \enquote{relative quality}) and unnormalized (representing \enquote{absolute quality}) features is desirable, which matches our prediction.

Finally, with respect to how the features are aggregated over the entire state, there seems to be a clear winner and it actually is the simplest one.
In the semantic labelling, there is a propositional formula that tracks the relationship between the subautomata (see \cite{sickert_meyer_philipp_modernizing_2021} for details).
We can aggregate the values alongside this formula ($\lor=\max, \land=\min$, $\neg=1-x$) to obtain a single value for the entire state.
We conjecture that this type of aggregation suffices for many cases whereas the more complex aggregation methods (e.g.\ feature A of the formula that maximizes feature B) are only useful in niche situations (but if they are useful, they are very useful).

\paragraph{Further Remarks and Summary}
In addition to the features discussed, there is one more type of feature employed by \emph{every} model we trained which is information on the edge priority.
This is not too surprising as the edge priorities alone are determining the winner of the game.
In combination with a feature measuring \enquote{formula-controllability} and one that measures \enquote{formula-sat-difficulty}, it marks the universally agreed upon feature-core of every one of our models.
This core allows for example for the implementation of a \enquote{follow winning edge priorities unless it leads to obviously lost states}-strategy, which we conjecture is roughly what happens in the model that uses only these three features.

In summary we can say that we built a solid spectrum of features, where several dimensions of the complexity of the semantic labelling are covered and which suffices to give meaningful exploration advice very consistently.
Nevertheless, we are confident that there are also many aspects of the semantic labelling that our current features do not cover and where further research is necessary to realize the full potential.}
{}

\end{document}